\documentclass{article}

\usepackage{microtype}
\usepackage{graphicx}
\usepackage{subcaption}
\usepackage{booktabs} 

\usepackage{hyperref}

\newcommand{\method}{GCOD }

\usepackage[preprint]{icml2026}

\usepackage{amsmath}
\usepackage{amssymb}
\usepackage{mathtools}
\usepackage{amsthm}

\usepackage{hyperref}
\usepackage{url}
\usepackage[utf8]{inputenc} 
\usepackage[T1]{fontenc}    
\usepackage{booktabs}       
\usepackage{amsfonts}       
\usepackage{nicefrac}       
\usepackage{microtype}      
\usepackage{xcolor}         
\usepackage{amsmath}
\usepackage{graphicx} 
\usepackage{natbib}
\usepackage{adjustbox}
\usepackage{multirow}
\usepackage{mathtools}
\usepackage{makecell}
\usepackage{amsthm}
\usepackage{bm}
\usepackage{subcaption}
\usepackage{xurl}

\usepackage[capitalize,noabbrev]{cleveref}

\theoremstyle{plain}
\newtheorem{theorem}{Theorem}[section]
\newtheorem{proposition}[theorem]{Proposition}

\theoremstyle{definition}

\theoremstyle{remark}
\newtheorem{remark}[theorem]{Remark}

\usepackage[textsize=tiny]{todonotes}

\icmltitlerunning{Energy Guided smoothness to improve Robustness in Graph Classification}

\begin{document}

\twocolumn[
  \icmltitle{Energy Guided smoothness to improve Robustness in Graph Classification}



  \icmlsetsymbol{equal}{*}

  \begin{icmlauthorlist}
    \icmlauthor{Farooq Ahmad Wani}{equal,yyy}
    \icmlauthor{Maria Sofia Bucarelli}{yyy,sch}
    \icmlauthor{Maria Vittoria Vestini}{yyy}
    \icmlauthor{Andrea Giuseppe Di Francesco}{yyy,pisa}
    \icmlauthor{Oleksandr Pryymak}{comp}
    \icmlauthor{Fabrizio Silvestri}{yyy}
  \end{icmlauthorlist}

  \icmlaffiliation{yyy}{Sapienza University of Rome, Rome, Italy}
  \icmlaffiliation{comp}{mindthehuman.ai, London, England}
  \icmlaffiliation{sch}{CNRS, i3S, Inria}
  \icmlaffiliation{pisa}{Institute of Information Science and Technologies "Alessandro Faedo" - ISTI-CNR, Pisa, Italy}
  
  \icmlcorrespondingauthor{Farooq Ahmad Wani}{farooqahmad.wani@uniroma1.it}

  \icmlkeywords{Machine Learning, ICML}

  \vskip 0.3in
]



\printAffiliationsAndNotice{}  
\begin{abstract}
Graph Neural Networks (GNNs) are vulnerable to label noise in graph classification, yet the mechanisms underlying this vulnerability remain poorly understood.
We discover that GNN robustness is governed by representation smoothness, as measured by Dirichlet energy: energy decreases while learning true patterns but sharply increases when memorizing noisy labels, particularly in high-frequency components.
Based on this insight, we propose three complementary methods---spectral weight constraints, explicit energy regularization, and a novel GCOD loss---that limit harmful high-frequency energy growth through different mechanisms.
Experiments across seven benchmarks under both symmetric and asymmetric noise demonstrate consistent improvements while preserving clean-data performance.
Together, these results establish Dirichlet energy as both a diagnostic signal for noise overfitting and a principled target for robust GNN design.
\end{abstract}


\section{Introduction}


%
Graph Neural Networks (GNNs) have become the dominant approach for learning on graph-structured data~\cite{dgcnn, diffpool, hgpsl}, with graph classification---assigning labels to entire graphs---finding applications from molecular property prediction~\cite{NCI} to protein function analysis~\cite{proteins} and social network classification~\cite{Yanardag2015a}. 
However, real-world datasets often contain label noise: molecules may be mislabeled due to experimental error, proteins may be incorrectly categorized, or social graphs may receive inconsistent annotations. 
While extensive work addresses label noise in image classification~\cite{algan2021image}, the graph domain presents unique challenges that remain underexplored.
Compared to image~\cite{algan2021image} or node classification~\cite{nrgnn11}, the problem of graph classification with noisy labels is relatively less explored.
While initial pioneering works \cite{nt2019revisiting, omg7} have begun to address this challenge, a systematic understanding of when and why GNNs are vulnerable to noisy labels and the development of robust mitigation strategies tailored to the uniquely address this challenge remain active areas of research.
In this paper, we address this noise robustness challenge and study GNNs' susceptibility to label noise when some samples are labeled incorrectly.
The conventional understanding is that cross entropy loss (CE), usually used in GNN classification tasks, typically leads to overfitting in the presence of noisy labels, particularly when the model has sufficient expressivity \cite{memorizenoise}.
We observed that, for graph classification, GNNs trained with standard CE can show varying degrees of robustness when exposed to label noise.
%
%
%
We hypothesize that the answer lies in the \emph{smoothness} of learned node representations.
GNNs aggregate information from neighboring nodes, naturally producing representations where connected nodes are similar---a property quantified by Dirichlet energy~\cite{dirichlet}. 
Since graph-level predictions are obtained by pooling these node representations, the smoothness bias propagates to the graph level.
We propose that this bias serves as implicit regularization: fitting a noisy graph label requires the pooled representation to differ from correctly-labeled graphs of the same class, which in turn requires some node representations to sharpen---a costly adaptation that manifests as increased Dirichlet energy.

Through empirical analysis, we confirm this link and show that noise overfitting in graph classification corresponds to node representations sharpening, as measured by rising Dirichlet energy.
Based on these insights, we develop methods to detect overfitting and propose three distinct strategies to enhance robustness in noisy graph classification.
Specifically, our contributions are the following:

\begin{itemize}
\item We present the first systematic study linking \textbf{label noise robustness in graph classification}  to the \textbf{spectral dynamics of Dirichlet energy ($E^{\text{dir}}$)}. While prior works have studied oversmoothing  and energy decay in node classification, we reveal how noise memorization in graph classification  corresponds to a characteristic rise in high frequency Dirichlet components.
\item We propose a \textbf{unifying energy based perspective} on robustness, showing that three seemingly 
different approaches (i) enforcing positive eigenvalues in GNN weights, (ii) directly regularizing 
Dirichlet energy, and (iii) introducing the novel GCOD loss can all be understood as mechanisms 
that constrain harmful high frequency energy.
\item We provide \textbf{comprehensive empirical evidence} across diverse benchmarks and both symmetric and asymmetric noise, establishing  Dirichlet energy as a reliable signal of overfitting. Crucially, our methods \textbf{improve robustness without degrading clean data performance}.
\end{itemize}

Together, these contributions introduce a principled framework that connects spectral smoothness, Dirichlet energy, and noise robust learning in GNNs. We believe this perspective opens a new direction for designing graph models that are not only robust to label noise, but also more stable under domain shift and adversarial perturbations. The code is available at \url{https://anonymous.4open.science/r/Robustness_Graph_Classification-E76F}.


\section{Related Works}
\label{sec:related_work}

\paragraph{Learning under label noise.}
A large body of work addresses learning with noisy labels in image classification. 
\textbf{Robust loss functions} use symmetric losses~\cite{ghosh2017robust} or loss correction~\cite{patrini2017making}. 
\textbf{Sample selection methods} exploit the early learning phenomenon---where networks fit clean patterns before memorizing noise~\cite{arpit2017closer}---to identify likely clean samples via small loss values~\cite{gui2021towards}. 
\textbf{Neighborhood-based approaches} detect outliers in feature space~\cite{zhu2022detecting}.
However, these methods were developed for images and do not exploit graph structure. Our GCOD loss adapts sample reweighting to graphs using graph-level representation centroids, while our energy-based methods leverage a graph-specific inductive bias: the smoothness of node representations induced by message passing.
Additional methods are discussed in Appendix~\ref{sec:learningundernoise}.

\paragraph{Graph learning under noise.}
Unlike images, graphs exhibit noise in labels, topology, and node features. 
Most prior work focuses on \emph{node classification}---labeling individual nodes within a single graph---where methods mainly exploit homophily (connected nodes sharing labels) to detect corrupted nodes~\cite{graphC1, contrastivelearning5, omg7, ssl8, cl9, nrgnn11, Kang2018RobustGL}. Noise at edge and feature levels has also been explored~\cite{structuralnoise2, structuralnoise3, ssl8}.
In contrast, \emph{graph classification} assigns a single label to an entire graph, where within-graph homophily/heterophily does not directly inform the graph-level label. 
Fewer studies address this setting: \cite{graphC1} propose a surrogate loss to discard noisy labels under restrictive assumptions, and \cite{omg7} combine contrastive learning with MixUp~\cite{mixup2} and curriculum learning to filter noisy samples.
Unlike these sample-selection approaches, we identify \emph{representation smoothness}---quantified by Dirichlet energy---as the mechanism governing robustness in graph classification, enabling both a diagnostic signal for overfitting and principled interventions. Our loss function is inspired by~\cite{wani2023combining} but adapted for graphs with energy-aware modifications. Additional discussion is in Appendix~\ref{sec:graphlearningundernoise}.

\paragraph{Dirichlet energy in GNNs.}
Dirichlet energy quantifies smoothness of node features across graph edges~\cite{dirichlet}. 
Most GNNs act as low-pass filters, causing energy to \emph{decrease} across layers~\cite{nt2019revisiting, rusch2023survey}. 
\cite{nt2019revisiting} showed this for graphs without non-trivial bipartite components. 
\cite{noteonoversmoothing, oversmoothing2} prove that energy decays exponentially with depth under certain spectral conditions, while \cite{GRAFF} show that weight matrix eigenvalues control smoothing versus sharpening.
This energy decay causes \textbf{oversmoothing}---a problem in \emph{node classification} where node representations collapse and become indistinguishable~\cite{2sides}. Various mitigation approaches exist~\cite{FAGCN, zhou2021dirichlet, EEConv}, all focused on node classification.

We reveal a complementary phenomenon in \emph{graph classification}: fitting noisy graph labels causes Dirichlet energy to \textbf{increase}, not decrease. 
Intuitively, misclassifying a graph requires its pooled representation to differ from correctly-labeled graphs, which necessitates sharp (high-frequency) node features.
While prior work addresses energy \emph{decay} harming node classification, we show that energy \emph{growth} signals noise memorization in graph classification---a novel connection enabling new robustness methods.
Comprehensive discussion is in Appendix~\ref{sec:dir_additional_related}.

\section{Background}
Let $\mathcal{G} = (\mathcal{V}, \mathcal{E}, \mathbf{X})$ be an undirected graph, with $\mathcal{V}$ the set of nodes and $\mathcal{E} $ the set of edges. We denote by $N = |\mathcal{V}| $  the number of nodes of   $\mathcal{G}$.  
$\mathcal{N}_u$ is the neighborhood of the node $u$, and $d_u = |\mathcal{N}_u|$ is its degree.
$\mathbf{D}$ $\in \mathbb{R}^{N \times N}$ is the degree matrix, a diagonal with  entries 
$D_{u u} = d_u$. Each node $u$ has feature vector
$\mathbf{x}_u \in \mathbb{R}^m$. The  feature
matrix  $\mathbf{X} \in \mathbb{R}^{N \times m} $ stacks all the feature vectors. $\mathbf{A} \in \{0, 1\}^{N \times N}$ is the graph's adjacency matrix, with $A_{uv} = 1$ if $(u, v) \in \mathcal{E}$  and $A_{uv} = 0$ otherwise.

\textbf{Graph Neural Networks for Graph Classification. }In graph classification, each sample in the dataset, $\mathcal{D}$, is a graph, i.e.,  $\mathcal{D} = \{\mathcal{G}^{i}, \mathbf{y}_i\}_{i = 1}^{n}$, where $\mathcal{G}^{i} = (\mathcal{V}^i, \mathcal{E}^i, \mathbf{X}^i)$, and $\mathbf{y}_i \in \{0, 1\}^{|C|}$ is its class associated one-hot encoded representation.
We represent the set of labels for $\mathcal{D}$ as $\mathbf{y} \in \{0, 1\}^{n \times |C|}$. More simply we use $c_i$ to express the class of sample $i$. $\mathbf{X}^i \in \mathbb{R}^{N_i \times m}$, and $\mathbf{A}^i \in \mathbb{R}^{N_i \times N_i}$ are the feature and adjacency matrices of graph $i$ respectively.
In the case of learning under label noise, in the training data $c_i$ may differ from the ground truth. GNNs are employed to extract features from graph structured data. 
In the message passing framework \cite{messagepassing}, each input feature matrix $\mathbf{X}^i = \mathbf{H}^0_i$ (for $i \in \{1, \ldots, n\}$) is iteratively updated through the GNN layers. At layer $l$ (where $0 \leq l \leq L$), the intermediate representation is denoted as $\mathbf{H}_i^l$, producing output latent features $\mathbf{Z}^i = \mathbf{H}^L_i \in \mathbb{R}^{N_i \times m'}$ for graph $\mathcal{G}^i$.
Given a set of weights $\mathbf{W}_l$ and $\mathbf{\Omega}_l$ for layer $l$, the message-passing update rule for graph $i$ is:
\begin{equation}
\begin{aligned}
\mathbf{H}^{l+1}_i
&=
UP_{\mathbf{\Omega}_l}\Big(
    \mathbf{H}^{l}_i,\;
    AGGR_{\mathbf{W}_l}\big(
        \mathbf{H}^{l}_i,\;
        \mathbf{A}^i
    \big)
\Big), \\
&\qquad 0 \leq l \leq L,\; l \in \mathbb{N}.
\end{aligned}
\end{equation}

where $UP_{\mathbf{\Omega}_l}$ and $AGGR_{\mathbf{W}_l}$ denote the \textit{update} and \textit{aggregation} functions of the message passing mechanism.
The final node representations \( \mathbf{Z}^i \in \mathbb{R}^{N_i \times m'} \), is then passed to a learnable, permutation-invariant function \( f_{\theta}: \mathbb{R}^{N_i \times m'} \rightarrow \mathbb{R}^{|C|} \) that transform them into class probabilities. The predicted output is then represented as a one hot encoded vector \( \mathbf{\hat{y}}_i \).


\textbf{Dirichlet Energy on graphs.} 
The Dirichlet energy for graph data: \( E^{dir} \) quantifies the smoothness of a scalar or vector field defined over the nodes of a graph. For a graph \( \mathcal{G}^i = (\mathcal{V}^i, \mathcal{E}^i) \) with node representation matrix \( \mathbf{Z}^i \in \mathbb{R}^{N_i \times m'} \), with \( \mathbf{Z}^i_u \) denoting the representation of node $u$, the Dirichlet energy is defined as:
\begin{equation}
E^{dir}(\mathbf{Z}^i) = \sum_{(u,v) \in \mathcal{E}^i} \left\| \mathbf{Z}^i_{u}/d_u - \mathbf{Z}^i_{v}/d_v \right\|_2^2
\label{eq:edir-edge}
\end{equation}
 Intuitively, $E^{dir}(\mathbf{Z}^i)$ is small when the connected nodes have similar representations (smooth signal), and large when the neighboring nodes differ (indicating sharpening).

\section{GNN Robustness to Noisy Graph Labels, and its Failures Modes}
\label{sec:failure}
\label{sec:gnn_robust_and_faulty_modes}

\begin{figure}[t]
\centering

\begin{subfigure}{0.47\columnwidth}
    \centering
    \includegraphics[width=\linewidth]{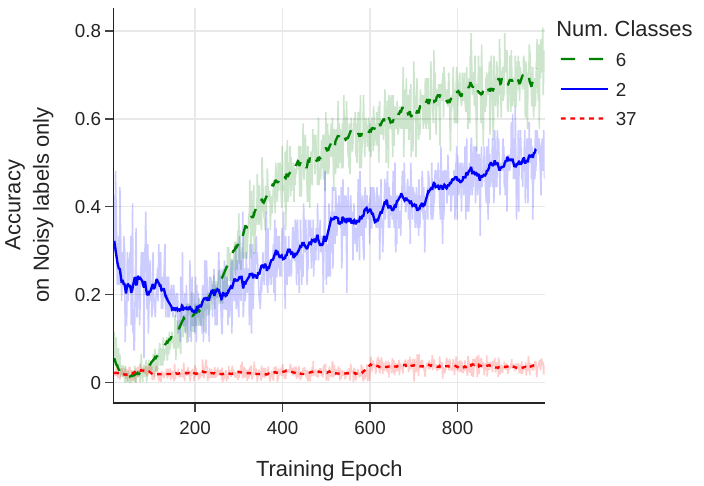}
    \caption{}
    \label{fig:ppa_changing_number_of_classes}
\end{subfigure}\hspace{0.01\columnwidth}
\begin{subfigure}{0.47\columnwidth}
    \centering
    \includegraphics[width=\linewidth]{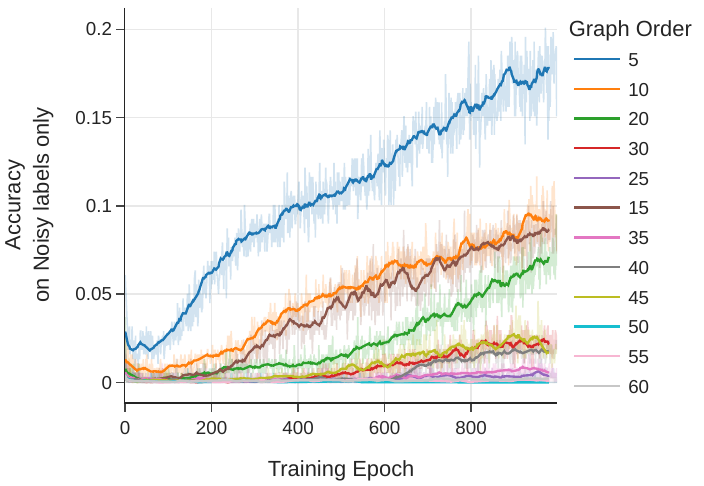}
    \caption{}
    \label{fig:synthetic_data}
\end{subfigure}

\caption{
Training accuracy on noisy labels only. Effect of dataset properties:
(a) Fewer classes in PPA lead to faster overfitting on noise.
(b) Lower graph order leads to faster overfitting on noise.
}
\label{fig:various_figures}
\end{figure}

\begin{figure}[t]
\centering

\begin{subfigure}{0.48\columnwidth}
    \includegraphics[width=\linewidth]{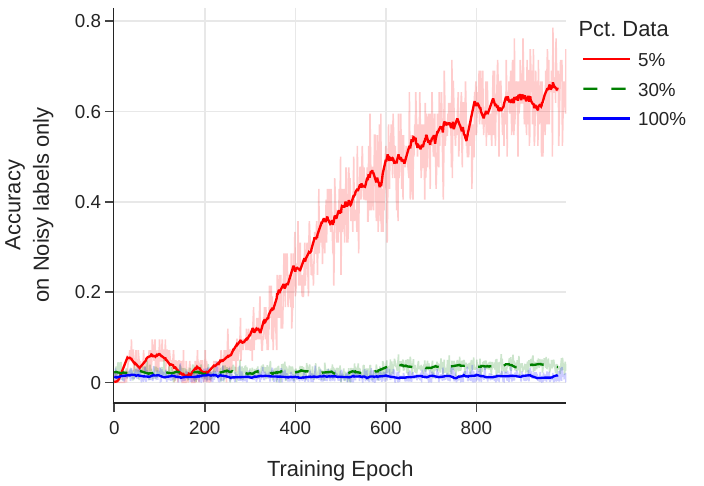}
    \caption{}
    \label{fig:ppa_changing_number_of_smaples_of_class}
\end{subfigure}
\hfill
\begin{subfigure}{0.48\columnwidth}
    \includegraphics[width=\linewidth]{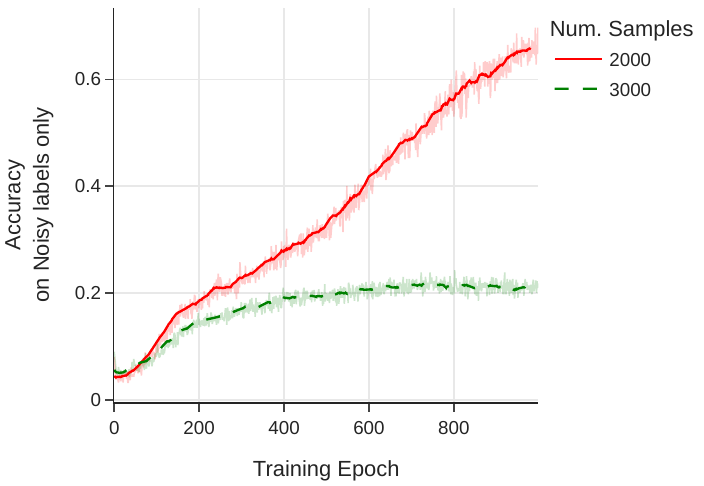}
    \caption{}
    \label{fig:varying_samples_synthetic}
\end{subfigure}

\caption{Training accuracy on noisy labels only. Effect of dataset size: (a) Smaller fractions of the PPA dataset lead to stronger noise memorization. (b) Smaller synthetic datasets are also more prone to memorizing noise.}
\end{figure}

Before developing robust methods, we must understand when and why standard GNNs fail under label noise.
This section systematically investigates failure modes through controlled experiments, revealing that vulnerability depends on the relationship between model capacity and task complexity.
We begin by noting that GNNs exhibit surprising robustness in some settings.
%
For instance, injecting noise into the full PPA  dataset ~\cite{PPA}, or large portions of it, does not significantly degrade model performance (see Fig.  \ref{fig:ppa_changing_number_of_classes}, \ref{fig:ppa_changing_number_of_smaples_of_class}, \ref{fig:noisy_pure_40_1}). 
We hypothesize that the observed robustness on PPA stems from the fact that the models may be under parameterized for the inherent difficulty of the PPA benchmark, on which state of the art methods struggle to achieve perfect accuracy\footnote{\url{https://paperswithcode.com/sota/graph-property-prediction-on-ogbg-ppa}}. 
This finding aligns with general findings that under parameterized models are often more robust to noise \cite{memorizenoise}.
Despite this, we show that GNNs nevertheless fail under certain conditions of label noise. 
To examine this, we fix the model architecture and manipulate task complexity by (i) varying the number of classes as a proxy for over-parameterization, (ii) varying the average number of nodes in synthetic datasets, and (iii) varying the share of training data used. 
We study the average training accuracy on noisy labels as a direct measure of how much the model fits noise. Higher accuracy on noisy samples indicates the model is memorizing them, while lower accuracy suggests that the model is not fitting the noise.

\textbf{GNN Robustness varying number of classes }
We use 30\% of the PPA dataset and inject 20\% symmetric noise into this subset by randomly replacing labels with uniformly sampled incorrect classes.
The model here and below, if not said otherwise, is a 5-layer Graph Isomorphism Network (GIN)~\cite{WL1} with 300 hidden units.
Specifically, we create a sub-sampled PPA dataset with 2, 6, and the full 37 classes. 
Intuitively, reducing the number of classes simplifies the classification task and reduces the effective dataset size, making the fixed model increasingly overparameterized relative to the task.
Fig.~\ref{fig:ppa_changing_number_of_classes} shows the training accuracy on noisy samples across epochs. 
For the full 37-class task, the GNN does not memorize noise and remains relatively robust. 
However, when the number of classes is reduced to 6 and further to 2, the model increasingly fits the noisy labels. 
Interestingly, the 2-class case exhibits slightly more robustness than the 6-class case due to the symmetric nature of the injected noise, since random flipping between two classes produces highly contrasting noisy samples.

\textbf{GNNs are not robust on low-order graphs.}
We generated synthetic datasets (see procedure in Appendix, Section \ref{appendix:details_synthetic_dataset_varying_nodes}) to study the effect of graph order (number of nodes). 
As shown in Fig.~\ref{fig:synthetic_data}, GNNs become increasingly sensitive to noise as the graph order decreases. 
Small graphs lack sufficient internal structure and aggregation capacity, making them vulnerable to treating noisy labels as signals.
Conversely, larger graphs provide more nodes over which the model can average, diluting the influence of noisy samples.

\textbf{GNNs are not robust on small training sets.}  
The size of a training set affects the robustness to noisy labels.
For the PPA dataset, we keep all 37 classes, but subsample the number of training graphs per class. 
As shown in Fig.~\ref{fig:ppa_changing_number_of_smaples_of_class}, reducing the number of training samples increases the likelihood of overfitting noise.
A similar trend is observed for the synthetic datasets (with graph order 7 and 6 classes) under 35\% label noise, as shown in Fig.~\ref {fig:varying_samples_synthetic}.
In both cases, models trained on smaller datasets have a higher tendency to memorize noisy samples due to insufficient clean data to learn generalizable patterns.

\section{Representation Dirichlet Energy indicates Overfitting on Noisy Labels}
\label{sec:Dirichlet_meet_GC}
\label{sec:dir_energy_fitting_noise}

\begin{figure}[t]
\centering

\begin{subfigure}{0.48\columnwidth}
    \includegraphics[width=\linewidth]{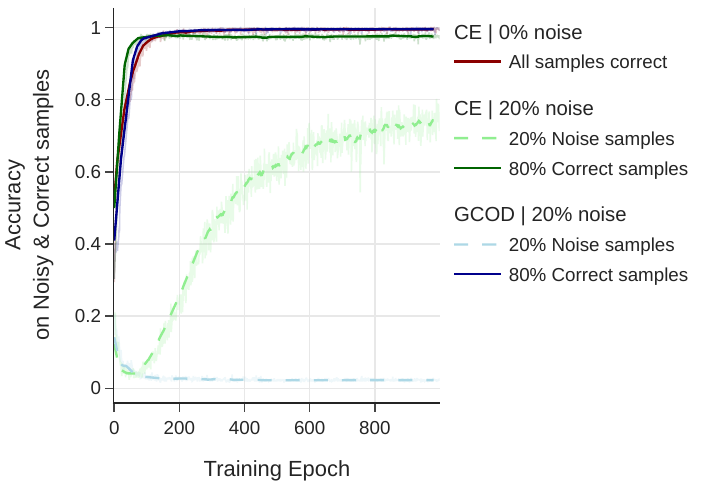}
    \caption{}
    \label{fig:PPA_DE_NCOD_subfig1}
\end{subfigure}
\hfill
\begin{subfigure}{0.48\columnwidth}
    \includegraphics[width=\linewidth]{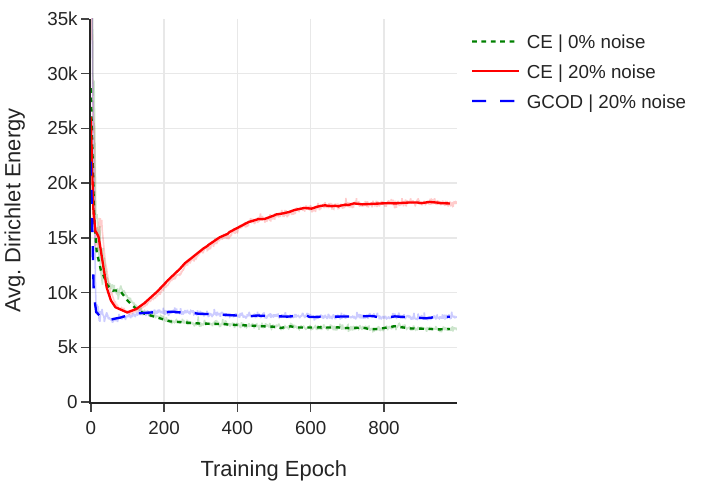}
    \caption{}
    \label{fig:PPA_NCOD_Dirich_energy}
\end{subfigure}

\caption{
(a) Evolution of training Accuracy for GIN model on the PPA dataset (30\% sample, 6 classes) with clean 0\% or 20\% label noise for CE and \method. 
(b) Dirichlet energy for clean and noise introduced PPA dataset (30\% sample, 6 classes). The Dirichlet energy increases when the model with CE fits on noise.
}
\label{fig:dirichlet_results}

\end{figure}

\begin{figure}[t]
\centering

\begin{subfigure}{0.48\columnwidth}
    \includegraphics[width=\linewidth]{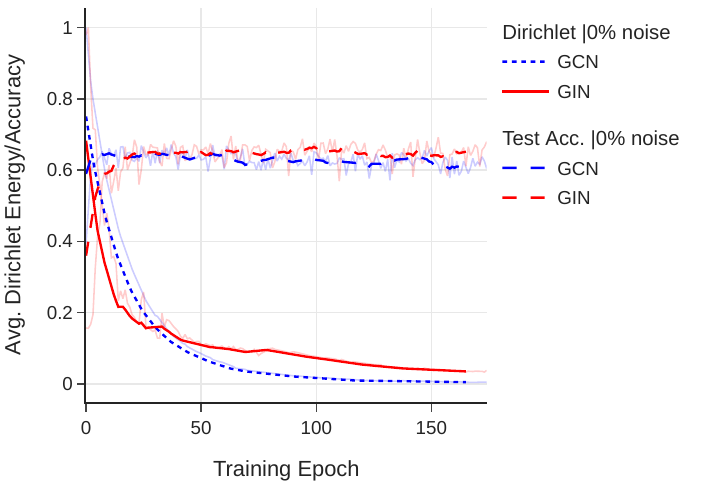}
    \caption{}
    \label{fig:gin_gcn_dir}
\end{subfigure}
\hfill
\begin{subfigure}{0.48\columnwidth}
    \includegraphics[width=\linewidth]{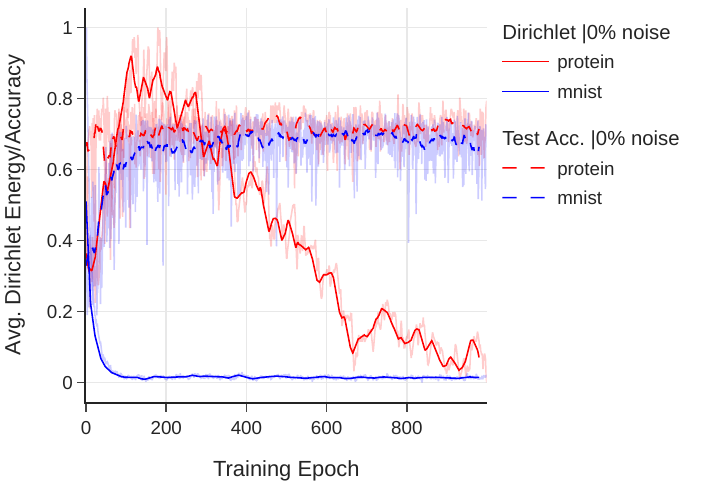}
    \caption{}
    \label{fig:comparedataset}
\end{subfigure}

\caption{
(c) Dirichlet energy and test Accuracy on the PPA dataset using CE with GIN and GCN models.
(d) Dirichlet energy and test Accuracy on different datasets using GIN model (axes scaled for comparison).
}
\label{fig:dircihlet_results1}

\end{figure}

Section~\ref{sec:failure} established that GNNs can overfit noisy labels when model capacity exceeds task requirements. We now investigate how this overfitting manifests in learned representations, revealing Dirichlet energy as a reliable diagnostic signal that also suggests intervention strategies.

Across all experimental setups, we consistently observe that the Dirichlet energy (\(E^{\text{dir}}\)) of the learned node representations increases once the model begins to fit on noisy labels (see Fig. \ref{fig:PPA_NCOD_Dirich_energy}). During early training, when the model captures true patterns, (\(E^{\text{dir}}\)) remains low; however, as the model starts fitting the noisy samples, the energy increases significantly. This consistent behavior across diverse datasets and conditions ( Fig. \ref{fig:PPA_NCOD_Dirich_energy} 
 and Figure \ref{fig:dirichlet_energy_comparison})  motivates us to use \(E^{\text{dir}}\) as a signal to detect and monitor overfitting to label noise. This leads to our first research question \textbf{RQ1}:\textit{ Is Dirichlet energy related to GNN performance on graph classification tasks, and how does it evolve when label noise is introduced during training?} 

To address this, we propose a graph-level Dirichlet energy measure for the graph classification task and analyze its empirical behavior during training under noise conditions.
Given a set of graphs $\mathcal{S}$, composed of graphs $\mathcal{G}^i$ with latent representation $\mathbf{Z}^i$, we define the Dirichlet energy of the set $\mathcal{S}$ as
\begin{equation}
    \label{eq: avg_dirichlet}
   E^{dir}_{\text{set}}(\mathcal{S}) := \frac{1}{|\mathcal{S}|} \sum_{\mathcal{G}^i \in \mathcal{S}}  E^{dir}(\mathbf{Z}^i).
\end{equation}
In particular, we study $E^{dir}_\text{set}(\mathcal{D}_c)$, the average Dirichlet energy of graphs belonging to class $c$, where $\mathcal{D}_c$ denotes the subset of training graphs with the label $c$. 

Our empirical analysis provides a consistent answer to \textbf{RQ1}. In clean datasets, i.e., without label noise, \( E^{dir}_{\text{set}}(\mathcal{D}_c) \) may fluctuate during the initial training phase as the model begins to adjust to the task. However, we consistently observe a steady decrease in the later stages of training, culminating in low and stable Dirichlet energy values once the model converges to high classification accuracy. This behavior is illustrated in Fig.~\ref{fig:gin_gcn_dir} and Fig. \ref{fig:comparedataset} for models trained with standard Cross-Entropy (CE) loss and no label noise (CE-0\%).
\looseness=-1

However, when synthetic label noise is introduced (e.g., 20\% symmetric flipping), the behavior diverges. As shown in Fig.~\ref{fig:PPA_DE_NCOD_subfig1}, while the initial phase of training still exhibits a decrease in \( E^{dir}_{\text{set}}(\mathcal{D}_c) \), this is followed by a significant increase during later epochs, precisely when the model starts to fit the noisy labels. This memorization phase is marked by rising training accuracy on noisy samples (see CE-20\%), demonstrating a direct link between noise fitting and increased Dirichlet energy.

This phenomenon is consistently observed across datasets and model architectures, including MUTAG, MNIST, and PROTEINS (see 
Fig. \ref{fig:comparedataset} and Fig. \ref{fig:mutag} in Appendix \ref{appendix:additional_figures}). These findings confirm that Dirichlet energy serves as a reliable signal of representation smoothness and its disruption as a result of noise memorization.
\paragraph{Why Smooth Representations Resist Noisy Graph Labels.}
This connection between smoothness and robustness admits an intuitive explanation specific to graph classification. 
In this setting, graph-level predictions are obtained by pooling node representations (e.g., via mean or sum pooling). 
For correctly labeled graphs, the GNN learns node representations that, when pooled, produce features characteristic of their class—and message passing naturally makes these representations smooth across neighbors.
Now consider a mislabeled graph: to fit this incorrect label, the pooled representation must resemble graphs from a \emph{different} class. 
Achieving this requires some node representations to deviate from the smooth patterns learned for correctly-labeled graphs—manifesting as sharp, high-frequency components that increase Dirichlet energy.
Thus, the smoothness bias inherent in GNN message-passing acts as implicit regularization against noisy \emph{graph} labels: the model can only fit mislabeled graphs by developing costly high-frequency representations, which it does only when capacity permits and training continues long enough.
This explains the U-shaped energy trajectory in Fig.~\ref{fig:PPA_NCOD_Dirich_energy}: energy decreases while learning true patterns, then increases when memorizing noise.

Furthermore, to isolate the spectral dynamics, we utilize the HLFF-GNN framework \cite{hlffgnn}, which decomposes the node representations into low-frequency \( \mathbf{Z}_1 \) and high-frequency \( \mathbf{Z}_2 \) components. Experiments show that while \( E^{dir}(\mathbf{Z}_1) \) remains stable, \( E^{dir}(\mathbf{Z}_2) \)( Appendix Figure~\ref{fig:dirichlet_energy_comparison}) sharply increases during noise overfitting, confirming that high frequency energy components are responsible for fitting mislabeled data (detailed analysis of these experiments is provided in Appendix \ref{sec:Amplification}).

From these observations, we conclude that maintaining a low Dirichlet energy, particularly by suppressing some high-frequency components, correlates with robust generalization. However, directly minimizing \( E^{dir}_{\text{set}}(\mathcal{S}) \) as a loss term presents practical challenges. First, the asymptotic energy level varies between datasets and architectures, making it difficult to define a universal target. Second, \( E^{dir}_{\text{set}} \) is a global dataset level quantity, which is not easily decomposed into sample gradients for stochastic optimization. We propose alternative strategies to promote smoothness.

\section{Robust Strategies Based on Smoothing}
\label{sec:two_smoothing_methods}
\subsection{Method 1: Robust GNN by enforcing positive Eigenvalues of transformations}
\label{sec:positive_eigens}

Our previous findings established a strong correlation between a GNN's overfitting of noisy labels and a significant increase in the Dirichlet energy of its learned node representations. 
The spectral properties of the learnable weight matrices within GNN layers fundamentally shape the network's behavior on the graph, particularly concerning smoothing and sharpening of features. Prior work~\cite{noteonoversmoothing, oversmoothing2, GRAFF} has shown that the eigenvalues of learnable weight matrices interact with the graph Laplacian, inducing either smoothing or sharpening effects. In particular, \cite{GRAFF} demonstrates that positive eigenvalues promote attraction between connected nodes, while negative eigenvalues induce repulsion.
These findings suggest that controlling the sign of the weight spectrum could architecturally enforce a smoothing inductive bias. 
This lead us to formulate our second research question: \textbf{RQ2} \textit{Does the spectrum of the weight matrices affect the evolution of \eqref{eq: avg_dirichlet} during training?}

To justify our dataset level analysis of Dirichlet energy, we first present the following result:

\begin{proposition}
\label{theorem: avg_dirichlet}
Let $\mathcal{D} = \{ \mathcal{G}^1= (\mathbf{Z}^1, \mathbf{A}^1), \ldots, \mathcal{G}^{n} = (\mathbf{Z}^n, \mathbf{A}^n) \}$ be a set of graphs. Then
\(
E^{dir}_\text{set}(\mathcal{D}) = \frac{1}{|\mathcal{D}|} E^{dir}(\mathbf{Z}),
\)
where $\mathbf{Z} = [\mathbf{Z^1} \| \ldots \| \mathbf{Z^n}]$ and $\mathbf{A}$ is a block-diagonal matrix with blocks $\mathbf{A}^i$ along the diagonal. That is, the dataset-level Dirichlet energy corresponds to the Dirichlet energy of a single disconnected graph composed of all graphs in $\mathcal{D}$.
\end{proposition} 
\begin{remark}
\label{cor: avg_energy}
Reducing $E^{dir}_\text{set}(\mathcal{D})$ during training implies that the model is simultaneously enhancing low-frequency representations across all graphs in the dataset.
\end{remark}

Discussion of Proposition~\ref{theorem: avg_dirichlet} and Remark~\ref{cor: avg_energy} is provided in Appendix~\ref{app: proof}. This result is particularly relevant for graph classification: it shows that controlling node-level smoothness across a dataset of graphs can be achieved through local spectral constraints, without requiring explicit graph-level regularization. Motivated by \cite{GRAFF}, we hypothesize that smoothing can be enhanced by removing negative eigenvalues from the learned weight matrices of the GNN. To test this hypothesis, we constrain the spectrum of the weight matrix \( \mathbf{W}^{(2)} \) in each GNN layer after neighborhood aggregation.
We refer to this approach as \textbf{CE+W2}, which uses standard cross entropy loss but with post-hoc positive eigenvalue enforcement on \( \mathbf{W}^{(2)} \).
\textit{The full derivation of the update rules, spectrum filtering, and implementation details are provided in Appendix~ \ref{app: weightmatrice_eigens}.} 
The performance of the method is reported in Table~\ref{tab:ppa_30_perc_6_classes}. 
Despite the findings affirm that controlling the weight matrix spectrum influences
Dirichlet energy and robustness, the eigen decomposition step introduces severe training overhead (Table \ref{tab:runtime_comparison}) and potential instability (Appendix Fig. \ref{fig:unstability}).


\subsection{Method 2: Robust GNN by direct energy manipulation}
\label{sec:regularization}

In this Section, We introduce a training method that explicitly constrains Dirichlet energy. By penalizing graphs with energy above a threshold, the model is encouraged to learn smoother, low-frequency representations, which we hypothesize improves robustness to label noise. Our approach is motivated by the empirical observations presented in Section \ref{sec:dir_energy_fitting_noise} and Appendix \ref{sec:Additional_studies}, which show that Dirichlet energy increases under overfitting and grows proportionally with label noise.

Formally, for a training set $\mathcal{D} = \{ \mathcal{G}_1, \ldots, \mathcal{G}_N \}$ with associated Dirichlet energies $E_i = E^{dir}(\mathcal{G}_i)$ and class labels $c_i$, we define the regularization term:
\begin{equation}
\label{eq: regularization term}
    \mathcal{L}_{DE}(\mathcal{D}) = \frac{1}{N} \sum_{i=1}^{N} \left[\max(0, E_i - U_{c_i})\right]^2,
\end{equation}
where $U_{c_i} \in \mathbb{R}$ is the energy threshold for class $c_i$. The overall training loss becomes \(
    \mathcal{L} = \mathcal{L}_{CE} + \lambda \mathcal{L}_{DE},\)
where $\lambda$ balances the smoothness constraint against the classification objective.
We explore two strategies for setting $U_c$ (see also Section~\ref{appendix:details_dirichlet_regularization_bounds} for more details on these strategies):\\
\textbf{Class-specific bound:} A dynamic threshold $U_c$ computed after each epoch as the average Dirichlet energy of clean validation graphs in class $c$. The validation set must be clean to provide a reliable reference for estimating class-dependent energy levels. When noise is high, especially symmetric noise, the class-specific upper bounds $U_c$ may lose discriminative power as energy distributions across classes become similar, reducing the method’s effectiveness. Using clean validation data preserves the class-specificity of the thresholds.\\
\textbf{Fixed bound:} A global threshold $U_c = U$ for all classes. In this case, the approach is not dynamic; instead $U$ is kept fixed during training and treated as a hyperparameter tuned to balance the need to prevent excessive smoothing while still limiting energy growth under noise.

On the PPA dataset with symmetric label noise, both variants of $\mathcal{L}_{DE}$ improved test accuracy over standard $\mathcal{L}_{CE}$. The fixed bound effectively constrained energy but occasionally over smoothed representations, particularly under clean labels. In contrast, class-specific bounds yielded better generalization and stability, improving accuracy under both noisy and clean conditions (Table~\ref{tab:ppa_30_perc_6_classes}). These results confirm that Dirichlet energy regularization helps stabilize feature evolution and enhances robustness by limiting harmful high frequency components.

\section{Method 3: Robust GNN with GCOD Loss Function}
\label{sec:GNNrobustnessLoss}

Having shown that noise overfitting aligns with increasing Dirichlet energy and that ($E^{dir}$) decreasing methods improve robustness, we now explore an alternative path: a robust loss function. We introduce Graph Centroid Outlier Discounting (\method), adapted from \cite{wani2023combining} for graph classification with noisy labels. Unlike previous methods, \method enhances robustness directly through its formulation, not by explicitly controlling $E^{dir}$.

While \method is not designed to directly minimize Dirichlet energy, we investigate its performance in the presence of label noise and observe the corresponding behavior in terms of $E^{dir}$. This leads to our research question: \textbf{RQ3}. \textit{Is \method able to prevent learning of noisy samples and promote smoothness of \eqref{eq: avg_dirichlet}, even though it is not specifically designed for it?}

\subsection{GCOD Loss Formulation}

NCOD~\cite{wani2023combining} is a loss function designed for image classification under label noise~\cite{memorizenoise}, which assumes samples of the same class cluster in latent space and exploits the tendency of networks to learn clean patterns before memorizing noise~\cite{arpit2017closer}.

We adapt this framework for graph classification. Let $f_{\theta}: \mathbb{R}^{N \times m'} \rightarrow \mathbb{R}^{|C|}$ map pooled node representations to class probabilities. For a batch of $B$ graphs, we introduce $\mathbf{u}_B \in \mathbb{R}^B$ as a learnable per-sample parameter capturing sample reliability, with $\mathbf{\hat{y}}_B$ (one-hot encoded) as predicted labels, $\mathbf{y}_B$ (one-hot encoded) as given labels, and $\mathbf{\tilde{y}}_B$ as soft labels computed from class centroids. Let $\mathbf{Z}_B$ denote the pooled graph representations for the batch.

\method optimizes three complementary terms:

\textbf{Term 1: Adaptive Cross-Entropy.}
\begin{equation}
\label{eq:gcod1}
\mathcal{L}_1 = \mathcal{L}_{\text{CE}}\big(f_{\theta}(\mathbf{Z}_B) + a_{\text{train}} \cdot \text{diag}_\text{vec}(\mathbf{u}_B) \cdot \mathbf{y}_B,\; \mathbf{\tilde{y}}_B\big)
\end{equation}
where $a_{\text{train}}$ is the current training accuracy. Early in training (low $a_{\text{train}}$), the loss behaves like standard CE; as training progresses, reliable samples (low $u$) are emphasized.

\textbf{Term 2: Reliability Learning.}
\begin{equation}
\label{eq:gcod2}
\mathcal{L}_2 = \frac{1}{|C|} \left\| \mathbf{\hat{y}}_B + \text{diag}_\text{vec}(\mathbf{u}_B) \cdot \mathbf{y}_B - \mathbf{y}_B \right\|^2
\end{equation}
This encourages $u_i$ to be small when graph $i$ aligns with its label (likely clean) and large when it deviates (likely noisy).

\textbf{Term 3: Alignment Regularization}:
\begin{equation}
\label{eq:gcod3}
\mathcal{L}_3 = (1 - a_{\text{train}}) \cdot D_{KL}\big(\mathfrak{L} \,\|\, \sigma(-\log(\mathbf{u}_B))\big)
\end{equation}
where $\mathfrak{L} = \log(\sigma(\text{diag}_\text{mat}(f_{\theta}(\mathbf{Z}_B)\mathbf{y}_B^\top)))$. This term, absent in the original NCOD, prevents $\mathbf{u}$ from growing aggressively early in training by regularizing alignment between predictions and reliability estimates.

\textbf{Parameter Updates.} Model parameters $\theta$ and reliability parameters $\mathbf{u}$ are updated separately:
\begin{equation}
\label{eq:gcod_update}
\theta^{t+1} \leftarrow \theta^t - \alpha \nabla_\theta (\mathcal{L}_1 + \mathcal{L}_3), \quad
\mathbf{u}^{t+1} \leftarrow \mathbf{u}^t - \beta \nabla_{\mathbf{u}} \mathcal{L}_2
\end{equation}

The soft labels $\mathbf{\tilde{y}}_B$ are computed from class centroids in representation space following~\cite{wani2023combining}; see Appendix~\ref{appendix:details_GCOD_method} for details.

\subsection{Connection to Graph Classification and Smoothness}

Unlike image classification where NCOD operates on flattened feature vectors, \method operates on \emph{pooled graph representations}---the result of aggregating node features across each graph. Graphs whose pooled representations deviate from their assigned class centroid are down-weighted, as they are likely mislabeled.

Although \method does not explicitly minimize Dirichlet energy, it achieves a similar effect. By down-weighting graphs that deviate from class centroids, \method prevents the model from learning sharp, high-frequency node patterns needed to fit outliers. Fig.~\ref{fig:PPA_DE_NCOD_subfig1} and~\ref{fig:PPA_NCOD_Dirich_energy} confirm this: under \method, Dirichlet energy remains low throughout training, whereas cross-entropy shows the characteristic energy spike during noise memorization.

\method consistently reduces overfitting on noisy samples and preserves smoothness in graph learning. Results for Graph Isomorphism Networks (GIN)~\cite{WL1} in Table~\ref{tab:ppa_30_perc_6_classes} and Table~\ref{tab:alldataset}, and for Graph Convolutional Networks (GCN)~\cite{gcn} in Fig.~\ref{fig:NCOD_accuracy} (Appendix), show that \method effectively mitigates noise impact, validating our hypothesis and answering \textbf{RQ3}.

\begin{table}[htbp]
  \centering

 
    \centering
    \caption{Performance on the PPA dataset using 30\% of the data restricted to 6 selected classes. The best test accuracy is highlighted in bold red, the second best in blue. Reported values denote mean ± standard deviation across 4 independent runs}
    \label{tab:ppa_30_perc_6_classes}

    \renewcommand{\arraystretch}{1.2}
    \resizebox{\linewidth}{!}{%
      \begin{tabular}{c c c c c c}
        \toprule
        \textbf{Noise} & \textbf{Method} &
        \multicolumn{2}{c}{\textbf{Test Acc.}} &
        \multicolumn{2}{c}{\textbf{Train Acc.}} \\
        \cline{3-6}
        & & \textbf{Best} & \textbf{Last} & \textbf{Best} & \textbf{Last} \\
        \midrule

        \multirow{5}{*}{0 \%}
          & CE              & 96.25 ± 0.05 & 91.25 &  1.00 & 99.33 \\
          & \method         & 96.65 ± 0.52 & 93.25 & 99.23 & 99.03 \\
          & CE + W2         & 96.50 ± 1.12 & 86.50 & 99.76 & 99.29 \\
          & Fixed           & 96.58 ± 0.27 & 85.70 & 99.26 & 98.29 \\
          & Class-specific  & 96.96 ± 0.04 & 88.67 & 99.41 & 98.67 \\
        \midrule

        \multirow{7}{*}{20 \%}
          & CE (clean only) & \textcolor{gray}{96.15 ± 0.09} & \textcolor{gray}{90.66}
                            & \textcolor{gray}{84.50} & \textcolor{gray}{84.07} \\ \cline{3-6}
          & CE              & 88.66 ± 0.16 & 62.58 & 94.98 & 93.45 \\
          & SOP             & 91.01 ± 0.44 & 85.50 & 79.17 & 77.64 \\
          & \textcolor{red}{\method}
                            & \textbf{\textcolor{red}{93.91 ± 0.26}} &
                              \textbf{\textcolor{red}{92.58}} &
                              \textcolor{red}{80.11} & \textcolor{red}{79.52} \\
          & CE + W2         & 89.83 ± 1.23 & 76.66 & 84.90 & 83.68 \\
          & Fixed           & 89.69 ± 0.57 & 80.25 & 78.07 & 70.14 \\
          & \textcolor{blue}{Class-specific}$^\ast$
                            & \textcolor{blue}{92.34 ± 0.41} &
                              \textcolor{blue}{80.92} &
                              \textcolor{blue}{84.55} & \textcolor{blue}{83.73} \\
        \midrule

        \multirow{7}{*}{40 \%}
          & CE (clean only) & \textcolor{gray}{95.08 ± 0.13} & \textcolor{gray}{80.44}
                            & \textcolor{gray}{68.15} & \textcolor{gray}{67.05} \\ \cline{3-6}
          & CE              & 82.08 ± 0.22 & 51.83 & 82.47 & 78.88 \\
          & SOP             & 82.33 ± 1.32 & 65.41 & 58.90 & 57.68 \\
          & \textcolor{red}{\method}
                            & \textbf{\textcolor{red}{93.88 ± 0.04}} &
                              \textbf{\textcolor{red}{91.08}} &
                              \textcolor{red}{65.09} & \textcolor{red}{64.31} \\
          & CE + W2         & 88.25 ± 1.13 & 56.33 & 68.78 & 65.88 \\
          & \textcolor{blue}{Fixed}
                            & \textcolor{blue}{88.58 ± 0.75} &
                              \textcolor{blue}{77.12} &
                              \textcolor{blue}{61.08} & \textcolor{blue}{54.53} \\
          & Class-specific$^\ast$
                            & 88.55 ± 0.11 & 71.83 & 68.98 & 67.80 \\
        \bottomrule
      \end{tabular}%
    }
    \end{table}
    \footnotetext{$^\ast$ \small Requires clean validation set.}

  \hfill
  \begin{table}
      \centering
     \caption{Performance of the GIN network across multiple datasets under 40\% asymmetric label noise. Reported values are test accuracy (\%). Columns correspond to cross-entropy (CE) with clean labels (0\% CE), cross entropy with noisy labels (40\% CE), and GCOD under 40\% noise (40\% GCOD).}
    \label{tab:asymmetric_noise}
     \resizebox{\linewidth}{!}{\begin{tabular}{lccc}
      \hline
      \textbf{Dataset} & \textbf{0\% CE} & \textbf{40\% CE} & \textbf{40\% GCOD} \\ \hline
      PROTEINS & 81.16 & 72.90 & 76.19 \\
      MNIST    & 72.69 & 71.15 & 72.61 \\
      ENZYMES  & 73.33 & 65.80 & 69.81 \\
      IMDB/B   & 76.50 & 68.18 & 72.89 \\
      MUTAG    & 94.73 & 91.16 & 93.19 \\
      REDDIT   & 48.15 & 48.01 & 47.94 \\
      MSRC/21  & 96.69 & 94.39 & 95.57 \\ \hline
    \end{tabular}}
  \end{table}

 \begin{table}
     \centering
     \centering
    \caption{Percentage improvement over cross entropy (CE) using the GIN network. Values show gains of OMG and GCOD on selected datasets.}
    \label{tab:omg_table}
    \begin{tabular}{lcc}
      \hline
      \textbf{Dataset} & \textbf{OMG} & \textbf{GCOD} \\ \hline
      MUTAG    & 0.061 & 0.062 \\
      IMDB-B   & 0.047 & 0.049 \\
      PROTEINS & 0.039 & 0.041 \\ \hline
    \end{tabular}    
 \end{table}

\section{Experimental Results}

We evaluate our three methods---spectral constraints (CE+W2), Dirichlet regularization ($\mathcal{L}_{DE}$ with fixed and class-specific bounds), and \method---against standard cross-entropy and recent robust baselines across diverse graph classification benchmarks under both symmetric and asymmetric noise. In addition to standard Cross Entropy (CE) baselines, we include two recent state-of-the-art methods: SOP~\cite{liu2022robusttraininglabelnoise}, a leading sample reweighting approach for noisy image classification, and OMG~\cite{omg7}, a graph-specific method combining contrastive learning with curriculum learning. Their inclusion provides strong comparative reference for evaluating noise robustness in graph classification.

\paragraph{Results on PPA}
Table \ref{tab:ppa_30_perc_6_classes} summarizes the results under 0\%, 20\%, and 40\% label noise on the PPA dataset, where 30\% of the data across $6$ classes was selected, utilizing a 5-layered GIN network (details provided in Appendix \ref{sec:experimental_setting}).

The proposed {GCOD} loss consistently outperforms other methods by achieving a smaller gap between the best and final accuracies, which reflects improved generalization and robustness to noise. 
SOP, despite being competitive, exhibits wider accuracy gaps, indicating its susceptibility to overfitting. 
{CE+W2} occasionally surpasses SOP at certain noise levels; its excessive smoothing leads to overfitting in noise-free scenarios. 
Precisely {CE+W2} improves test accuracy under moderate noise (e.g., 20\%: 89.83 vs. 88.66 compared to CE) and narrows the gap between training and test performance, indicating reduced overfitting. 
However, under clean labels (0\% noise), CE+W2 tends to over-smooth, with slightly degraded final accuracy. The eigen decomposition step introduces a modest training overhead (see Table \ref{tab:runtime_comparison}) and potential instability. However, the findings confirm that controlling the weight matrix spectrum influences Dirichlet energy and robustness. 

$\mathcal L_{DE}$ with a \emph{Fixed bound} shows results in line with CE+W2: it improves performance in the presence of moderate levels of noise, but in noise-free settings the gains were limited due to potential over-smoothing. With the use of \emph{Class-specific bounds}, instead, the adaptive mechanism allowed the regularization to align with the intrinsic complexity of each class, enabling the method to improve accuracy at all levels of label noise, including the absence of noise. This suggests the class-specific method improves generalization as well as model robustness.

\textbf{Results Across Multiple Datasets (20\% Symmetric Noise)}
Table \ref{tab:alldataset} compares {GCOD} with standard CE under 20\% symmetrical noise across several datasets.
The Table shows that {GCOD} outperforms CE on most datasets, highlighting its resilience regardless of the specific data characteristics (with the exception of REDDIT-MULTI-12K in this specific test).

\textbf{Results under Asymmetric Noise (40\%)}
In Table~\ref{tab:asymmetric_noise}, a comparison of {GCOD} and CE under 40\% asymmetric label noise across datasets is presented. {GCOD} consistently outperforms CE, demonstrating its robustness in handling asymmetric label noise.

\textbf{Comparison with OMG}
Lastly, Table~\ref{tab:omg_table} compares the percentage accuracy improvements of the OMG method and our proposed {GCOD} method across three datasets (MUTAG, IMDB-B, and PROTEINS) under experimental conditions similar to those in the OMG paper. It further underscores the superior performance and robustness of {GCOD} in noisy environments. In terms of efficiency, Table IV of the OMG paper reports an average runtime overhead of ~200\% for small datasets (MUTAG, PROTEINS, IMDB-B, NCI1) compared to a base GIN. OMG is even slower than Co-Teaching under identical settings. In contrast, GCOD has only a 0.029\% overhead even on the large and complex OGB-PPA dataset.

\textbf{Computational Efficiency and Hyperparameter Sensitivity of GCOD} 
Table~\ref{tab:runtime_comparison} compares the percentage runtime increase for various methods relative to GIN trained with cross entropy loss, normalized to 1. {CE+W2} incurs a 33\% increase in training time. The GCOD loss function introduces no additional hyperparameters beyond the learning rate for the learnable parameters (the weights and a parameter $u$). Table~\ref{tab:lr_sensitivity} shows the impact of the learning rate of $u$ on GCOD performances.

\subsection{Evidence for Causality}

A natural concern is whether Dirichlet energy merely \emph{correlates} with noise robustness or \emph{causally} influences it. Three observations support a causal relationship: (i)~\textbf{Convergent mechanisms}---our three methods improve robustness through fundamentally different mechanisms (spectral constraints, explicit energy penalties, centroid-based reweighting), yet all reduce Dirichlet energy under noise (Fig.~\ref{fig:PPA_NCOD_Dirich_energy}); (ii)~\textbf{Dose-response}---higher noise levels produce proportionally larger energy increases (Fig.~\ref{fig:noise_levels_energy}, Appendix~\ref{sec:Additional_studies}); and (iii)~\textbf{Temporal alignment}---energy spikes precisely when training accuracy on noisy samples rises (Fig.~\ref{fig:PPA_DE_NCOD_subfig1}), not gradually throughout training. While formal causal analysis is beyond our scope, these observations collectively support that controlling Dirichlet energy is a \emph{mechanism} for robustness, not merely a correlate.

\section{Conclusions}

We established Dirichlet energy as both a diagnostic signal for noise overfitting and a principled intervention target in graph classification. 
Our key finding—that noise memorization manifests as energy \emph{increase} rather than the energy decay seen in node classification—suggests that the smoothness bias of GNNs provides implicit robustness that practitioners can leverage.
Among our three methods, we recommend \method for general use: it achieves the strongest results, requires no clean validation data, and adds negligible overhead.
For practitioners who can access clean validation samples, class-specific $\mathcal{L}_{DE}$  offers an interpretable alternative with explicit energy control.
Beyond noisy labels, Dirichlet energy may serve as a general diagnostic for GNN training dynamics. 
We anticipate applications to domain shift detection, adversarial robustness, and architecture design.
\textbf{Limitations.} Our analysis is primarily empirical; theoretical characterization of when and why noise induces sharpening remains open. 
The class-specific regularization variant requires clean validation data, limiting its applicability in fully noisy settings.
%
%

\section*{Broader Impact}

This work addresses robustness to label noise in graph classification, relevant to many fields like molecular screening, protein analysis, and network classification. 

\textbf{Positive impacts.} Improved robustness could reduce costly false predictions in drug discovery pipelines and enable reliable learning from imperfectly curated scientific datasets.

\textbf{Potential concerns.} Graph classification methods, including ours, could in principle be applied to social network analysis for purposes such as user profiling. However, our contribution—robustness to label noise—does not expand the capability frontier of such applications; it merely improves reliability of existing methods.

\section*{Acknowledgments}
Maria Sofia Bucarelli has been supported by the French government, through the 3IA Cote d’Azur Investments in the project managed by the National Research Agency (ANR) with the reference number ANR-23-IACL-0001. Farooq Ahmad Wani was supported by the PNRR MUR project IR0000013 – SoBigData.it.



\bibliography{main}
\bibliographystyle{icml2026}

\newpage
\appendix

\onecolumn 
\section{Appendix}
This appendix provides additional details, extended experiments, and proofs that support the main paper. Section~A describes the experimental settings and datasets. Section~B presents further details on the proposed methods. Section~C reviews additional related works. Section~D–H contain extended experiments, ablation studies, and supplementary figures.
\section{ Experimental Settings} 
\label{sec:experimental_setting}
{
In our experimental setup, we performed tests on several datasets to evaluate performance. The first dataset, ogbg-ppa \cite{PPA}, consists of undirected protein association neighborhoods derived from the protein-protein association networks of 1,581 species, spanning 37 broad taxonomic groups. Another dataset, ENZYMES \cite{proteins}, includes 600 protein tertiary structures from the BRENDA enzyme database, representing six different enzyme classes. The MSRC\_21 dataset \cite{Neumann2016} contains 563 graphs across 20 categories, with an average of 77.52 nodes per graph. The PROTEINS \cite{proteins} dataset is a binary classification set with 1,113 graphs, having an average node count of 39.06 per graph. The MUTAG \cite{mutag} dataset is another small binary graph dataset, consisting of 188 graphs, each with an average of 18 nodes. The IMDB-BINARY \cite{imdbbinary} dataset, as the name suggests, is a binary graph classification dataset containing 1,000 graphs with an average node count of 20 per graph, and no node features. Similarly, the REDDIT-MULTI-12K dataset \cite{imdbbinary} includes 11,929 graphs spread across 11 classes, with an average of 391 nodes per graph and no node features.

Additionally, we utilized the MNIST graph dataset, which is derived from the MNIST computer vision dataset. This dataset contains 55,000 images divided into 10 classes, where each image is represented as a graph. 

Our experimental investigations were primarily conducted employing Graph Convolutional Networks (GCN)  \cite{gcn} and Graph Isomorphism Networks (GIN) \cite{WL1} networks. Notably, the experimental methodology adopted possesses a generality that extends to encompass all Message Passing Neural Networks (MPNNs). Our study centers on observing the learning dynamics of networks during graph classification, particularly examining their adaptability to label noise. We aim to enhance robustness by employing tailored loss functions. Notably, the selection of hyperparameters remains unrestricted, as these parameters depend on both the model architecture and the dataset employed, ensuring a nuanced and generalized approach. In each experiment, we initialize with hyperparameters suited for clean, non-noisy conditions, ensuring optimal model performance. These parameters are subsequently held constant as we introduce varying levels of noise, sample density, or graph order. This approach ensures fair comparisons across experiments and facilitates a comprehensive exploration of model capacities. The synthetic label noise is generated following the methodologies described in \cite{https://doi.org/10.48550/arxiv.1804.06872} and \cite{xia2021robust} which are considered to be standard techniques for generating synthetic label noise.


\subsection{Hyperparameters}
We employed the standard GIN \cite{WL1} and GCN \cite{gcn} architectures for most of our experiments. However, to investigate the impact of positive eigenvalues on weight matrices, as outlined in \ref{sec:positive_eigens}, we applied targeted modifications to both the GIN and GCN models \ref{app: weightmatrice_eigens}.

The table below summarizes the key hyperparameters used for the experiments.

\begin{table}[h!]
\centering
\begin{tabular}{|c|c|c|}
\hline
\textbf{Parameter}        & \textbf{Value}                                    \\ \hline
\textbf{Architecture}     & GCN, GIN                                     \\ \hline
\textbf{Learning Rate}    & 0.001                                     \\ \hline
\textbf{Optimizer}        & Adam,                                    \\ \hline
\textbf{Batch Size}       & 32                                               \\ \hline
\textbf{Loss Function}    & CrossEntropy, SOP and GCOD                 \\ \hline
\textbf{Epochs}           & 200(PPA) to 1000                                              \\ \hline
\textbf{Noise Percentage} & 0\% to 40\%                   \\ \hline
\textbf{Weight Decay}     & 1e-4                       \\ \hline
\textbf{Evaluation Metric} & Accuracy                \\ \hline
\textbf{GNN Layers } & 5                \\ \hline
\textbf{Learning Rate $u$ } & 1               \\ \hline
\textbf{Hidden Units } & 300             \\ \hline
\end{tabular}
\caption{Network architecture and hyperparameters.}
\end{table}

All experiments have been performed over NVIDIA RTX A6000 GPU. For implementation, visit the following anonymous GitHub: 
\href{https://anonymous.4open.science/r/Robustness_Graph_Classification-E76F}{Robustness Graph Classification Project}.

\subsection{Details on the synthetic dataset used for Figure~\ref{fig:synthetic_data} } \label{appendix:details_synthetic_dataset_varying_nodes}
Figure~\ref{fig:synthetic_data} presents results from synthetic datasets with varying graph order (i.e., number of nodes per graph). We generate datasets with average graph orders ranging from 5 to 60, sampling actual node counts from a Poisson distribution with mean equal to the target graph order. Each dataset contains 6 classes, with 1400 graphs per class, a fixed average degree of 2, and edges sampled uniformly at random.

Node features are sampled from a Gaussian distribution with a mean determined by the class label and a standard deviation of 1.5. For all graph orders, we apply a consistent label noise rate of 35\% using uniform class flipping. Results demonstrate that GNNs become increasingly sensitive to noise as the graph order decreases. Small graphs lack sufficient internal structure and aggregation capacity, making them vulnerable to treating noisy labels as signal. Conversely, larger graphs provide more nodes and connectivity over which the model can average, diluting the influence of noisy samples.

\section{Additional Details on the three proposed Methods}

\subsection{Spectral Bias Implementation Details}
\label{app: weightmatrice_eigens}
\subsubsection{GNN Update Rules and Spectral Analysis Setup}

To operationalize spectral bias in GNNs, we evaluate both GCN~\cite{gcn} and GIN~\cite{WL1} layer update mechanisms. For each input graph $\mathcal{G}_i$, we define the GCN update rule as:

\begin{equation}
    \text{GCN: \hspace{0.5cm}}
    \mathbf{H}_i^{l+1} = \sigma(\mathbf{\Delta} \mathbf{H}_i^l \mathbf{W}^1_l)\mathbf{W}^2_l, \quad \forall i \in \{1, \ldots, n\},
\end{equation}

and the GIN update rule as:

\begin{equation}
\begin{aligned}
    \text{GIN: \hspace{0.5cm}} \mathbf{H}_i^{l+1} = \sigma\left( \sigma\left((1 + \epsilon)\mathbf{H}_i^l + \mathbf{A}_i \mathbf{H}_i^l \right) \mathbf{W}_l^1 \right) \mathbf{W}_l^2, \\
    \forall i \in \{1, \ldots, n\}.
\end{aligned}
\end{equation}

Here, $\epsilon$ is a scalar hyperparameter and both $\mathbf{W}_l^1$ and $\mathbf{W}_l^2$ are square matrices of size $\mathbb{R}^{m' \times m'}$, allowing direct eigendecomposition. $\mathbf{W}_l^2$ is used as a shallow projection matrix after message aggregation.

\subsection{ Weight Matrix Spectrum and Clipping Procedure}
\label{app: eigendecomp}
For each layer $l$ and each weight matrix $v \in \{1, 2\}$, let the eigenvalues and eigenvectors be denoted:
\[
\{\mu_{0,l}^{v}, \ldots, \mu_{m'-1,l}^{v} \}, \quad \{\bm{\Phi}_{0,l}^{v}, \ldots, \bm{\Phi}_{m'-1,l}^{v} \}.
\]

Unlike the graph Laplacian $\mathbf{\Delta}$, these eigenvalues $\mu_{i,l}^v$ can be negative, which enables feature "sharpening" effects. As shown in \cite{GRAFF}, weight matrices with negative eigenvalues can amplify high frequency components often associated with noisy or irregular node signals.

To counteract this, we enforce a spectral bias by eliminating the influence of negative eigenvalues. The procedure is as follows:
\begin{enumerate}
    \item Compute the eigendecomposition:
    \[
    \mathbf{W}_l^v = \bm{\Phi}_l^v \bm{\mu}_l^v (\bm{\Phi}_l^{v})^{-1},
    \]
    where $\bm{\mu}_l^v$ is a diagonal matrix of eigenvalues.
    
    \item Apply element-wise ReLU to retain only non-negative eigenvalues:
    \[
    \bm{\mu}_l^{v+} = [\bm{\mu}_l^v]^+ = \max(\bm{\mu}_l^v, 0).
    \]

    \item Reconstruct the filtered weight matrix:
    \[
    \mathbf{W}_l^{v+} = \bm{\Phi}_l^v \bm{\mu}_l^{v+} (\bm{\Phi}_l^{v})^{-1}.
    \]
\end{enumerate}

This process removes sharpening components from the learned transformations, effectively biasing the GNN towards smooth solutions.

\subsubsection{ Training Integration and Backpropagation Handling}

In practice, we apply this spectral projection **after each gradient update**, treating it as a deterministic architectural constraint rather than part of the loss. The operation is not included in the computational graph—no gradients are propagated through the eigendecomposition or clipping.

This ensures that the model learns using unconstrained gradients, but the actual transformation used in forward passes remains positive-semidefinite.

\subsection{Detail on the Design of our \method }
\label{appendix:details_GCOD_method}
In our notation, $f_{\theta}: \mathbb{R}^{N \times m'} \rightarrow \mathbb{R}^{|C|}$ maps the final node representations \(\mathbf{Z}\in\mathbb{R}^{N \times m'}\) to the class probabilities.
We apply $f_{\theta}$ to batches of size $ \mathbf{B}$, and introduce $\mathbf{u}_B \in \mathbb{R}^B$ as a trainable parameter, with \(\mathbf{\hat{y}}_B \) as one-hot encoded class predictions, and $\mathbf{\tilde{y}}_B$ as calculated soft labels and as in \cite{wani2023combining}.

The \( \mathbf{Z_B}\) is the tensor containing node representation for each graph in the batch, 
$\text{diag}_{\text{mat}}(M)$  Extracting the diagonal elements of a matrix $M$, while 
$\text{diag}_{\text{vec}}(\mathbf{v})$ construct  a diagonal matrix from a vector $\mathbf{v}$. Here we offer its extension to Graph tasks with a new \method:

\begin{align}
\label{eq: eq_L1}
& \mathcal{L}_1(\mathbf{u}_B, f_{\theta}(\mathbf{Z}_B), \mathbf{y}_B, \mathbf{\tilde{y}}_B, a_{\text{train}}) = \mathcal{L}_{\text{CE}}( f_{\theta}(\mathbf{Z}_B) + a_{\text{train}} \text{diag}_\text{vec}(\mathbf{u}_B) \cdot \mathbf{y}_B, \mathbf{\tilde{y}}_B),  \\
\label{eq: eq_L2} 
& \mathcal{L}_2(\mathbf{u}_B, \mathbf{\hat{y}}_B, \mathbf{y}_B)  = \frac{1}{|C|} \left\| \mathbf{\hat{y}}_B + \text{diag}_\text{vec}(\mathbf{u}_B) \cdot \mathbf{y}_B - \mathbf{y}_B \right\|^2, \\
\label{eq: eq_L3}
& \mathcal{L}_3(\mathbf{u}_B, f_{\theta}(\mathbf{Z}_B), \mathbf{y}_B, a_{\text{train}}) = (1 - a_{\text{train}}) \mathcal{D}_{KL} \left\{ \mathfrak{L}, \sigma\left(-\log\left(\mathbf{u}_B\right)\right) \right\}
\end{align}

where  $\mathfrak{L}$ is $\log(\sigma \left( \text{diag}_\text{mat}(f_{\theta}(\mathbf{Z}_B)\mathbf{{y}_B}^T) \right))$ and $a_{\text{train}}$ is training accuracy. 

Equation \ref{eq: eq_L3}, is an additional term w.r.t vanilla NCOD, where we employ the Kullback-Liebler divergence as a regularization term to regulate the alignment of model predictions with the true class for clean samples (small $u$) while preventing alignment for noisy samples (large $u$). Moreover in \eqref{eq: eq_L1}, \ref{eq: eq_L2}, we insert $a_{train}$ as a feedback term.\\

The parameters of the losses are updated using stochastic gradient descent as follows:
\begin{align}
\mathbf{\theta}^{t+1} \gets \mathbf{\theta}^t - \alpha \nabla_\theta (\mathcal{L}_1 + \mathcal{L}_3) \qquad \qquad \mathbf{u}^{t+1} \gets \mathbf{u}^t - \beta \nabla_u \mathcal{L}_2
\end{align}
The parameter \( \mathbf{u} \) helps to reduce the importance of noisy labels during training, allowing the model to focus more on clean data. The computation of the soft label $\mathbf{\tilde{y}}_i \in \mathbb{R}^{|C|}$ (i.e. the $i$-th row of $\mathbf{\tilde{y}}$) relies on the concept of class embedding \cite{wani2023combining}. 

\subsubsection{Ablation: Training Accuracy Feedback}

The incorporation of $a_{\text{train}}$ in Equations~\ref{eq:gcod1} and~\ref{eq:gcod3} is critical. Without $a_{\text{train}}$ modulation, $\mathbf{u}$ dominates early in training, prematurely discounting samples before meaningful patterns are learned. With $a_{\text{train}}$, discounting activates gradually, yielding 3--5\% higher test accuracy under 20\% noise.

\subsection{Details on the Definition of the Bounds for $\mathcal L_{DE}$}
\label{appendix:details_dirichlet_regularization_bounds}
We defined two strategies for defining the upper bound for the regularization term $\mathcal L_{DE}$: a fixed global threshold and a class-dependent adaptive threshold.

\textbf{Class-dependent.} This approach is motivated by the observation that Dirichlet energy is influenced by factors that can be inherent to each class, such as graph topology. As a result, graphs from different classes may naturally exhibit distinct Dirichlet energy distributions.

To address this variability, as stated in the main text, we proposed the class-specific bound formulation. For each class $c$, the upper bound $U_c$ is computed at each epoch as the average Dirichlet energy over the clean validation graphs belonging to class $c$. Formally, given $\mathcal D^{val}_c$ the set of validation graphs in class c, and $E_i$ the Dirichlet energy of graph $\mathcal G_i$, the bound $U_c$ is computed as:
\begin{equation}
    U_c = \frac 1 {|\mathcal D^{val}_c|} \sum_{\mathcal G_i \in \mathcal D^{val}_c} E_i
\end{equation}

The use of clean validation data is essential for ensuring that the thresholds $U_c$ are reliable indicators of the intrinsic smoothness or complexity associated with each class. Relying on noisy samples to compute $U_c$, especially in the case of high symmetric label noise, would distort the energy, causing different class thresholds to collapse toward similar values. This would reduce the discriminative power of the regularization and lead $U_c$ to not be reflective of the true underlying structure of each class. \\
This adaptive strategy, then, ensures the regularization remains sensitive to plausible class-dependent variations between the distributions of the energy, which prevents the over penalization of inherently complex classes and under penalization of simpler ones.

\textbf{Fixed.} For the fixed settings, a constant threshold $U$ was applied uniformly across all the training samples, regardless of the class. This approach simplifies the regularization term and, by enforcing uniform penalization, provides a consistent regularization framework.. 

However, careful tuning of $U$ was necessary. If set too high, the regularization effect is negligible, allowing the model to overfit noise; if set too low, excessive smoothing occurs, causing a notable drop in accuracy due to the model’s reduced ability to capture and distinguish important variations in the data. Consequently, $U$ was progressively decreased during experimentation until such a performance drop became evident. The selected value of $U$  thus represents a trade-off: energy is sufficiently reduced to prevent overfitting on noisy labels, while maintaining the model's capacity to distinguish between classes.

}


\section{Additional Related Works}
\label{sec:additional_related}
\subsection{Learning under label noise.}
\label{sec:learningundernoise}
Some methods focus on sample relabelling \cite{arazo2019unsupervised, reed2014training}.
Another family of techniques address noisy labels using two networks, splitting the training set and training two models for mutual assessment \cite{han2018co, li2020dividemix, kim2023crosssplit}. 
Regarding \textbf{regularization} for noisy labels, mixup augmentation \cite{mixup} is a widely used method that generates extra instances through linear interpolation between pairs of samples in both image and label spaces. Additionally, exist also \textbf{Reweighting techniques} aiming to improve the quality of training data by using adaptive weights in the loss for each sample \cite{liu2015classification, pleiss2020identifying}.

\subsection{Graph Learning in noisy scenarios.} 
\label{sec:graphlearningundernoise}
Works on node classification under label noise attempt to learn to predict the correct node label when a certain proportion of labels of the graph nodes are corrupted. 
In \cite{pairwiseinteraction4}, authors exploit the pairwise interactions existing among nodes to regularize the classification.
Other approaches use regularizes that detecting those nodes that are associated with the wrong information. Among these are contrastive losses \cite{contrastivelearning5, cl9}, to mitigate the impact of a false supervised signal. Then in \cite{ssl8}, it was also proposed a self supervised learning method to produce pseudo labels assigned to each node. Other mechanisms that employ pseudo-labels are discussed in \cite{noisegovernance6}, showing different policies to down weight the effect of noisy candidates into the final loss function.

The parallel line of work concerning GNN under noise is related to noise coming from missing or additional edges, and also noisy features. In \cite{structuralnoise2}, they focus on structural noise. They show that adding edges to the graph degrades the performance of the architecture. And propose a node augmentation strategy that repairs the performance degradation. However, this method is only tested with synthetic graphs. In \cite{structuralnoise3}, they develop a robust GNN for both noisy graphs and label sparsity issues (RS-GNN). Specifically, they simultaneously tackle the two issues by learning a link predictor that down weights noisy edges, so as to connect nodes with high similarity and facilitate the message passing. RS-GNN uses a link predictor instead of direct graph learning to save computational cost.
The link predictor is MLP-based since edges can be corrupted. Their assumption is that node features of adjacent nodes will be similar. Once the dense adjacency matrix is reconstructed it is used to classify nodes through GCN. Even though these methods achieve state of the art performance they are specifically designed for node classification and have some assumptions on the input graph, such as the homophily property \cite{structuralnoise3, pairwiseinteraction4, contrastivelearning5, nrgnn11}. Moreover, some of these are validated only within graphs with the same semantics \cite{structuralnoise3, pairwiseinteraction4, ssl8, cl9, nrgnn11} (e.g. citation networks), where the homophily assumption could be valid, but limiting for the overall research impact.

\subsection{Dirichlet Energy.}
\label{sec:dir_additional_related}
Graph Neural Networks (GNNs) face several challenges, including limited message passing expressiveness \cite{WL2}, over smoothing \cite{oversmoothing2}, and over-squashing \cite{oversquashing3}. Over-smoothing has been studied using Dirichlet energy \cite{dirichlet}, which quantifies signal smoothness across graph nodes. Previous research explores the relationship between energy evolution and over smoothing \cite{noteonoversmoothing, nt2019revisiting}, highlighting design choices that exacerbate this issue. Various approaches have been proposed to mitigate over-smoothing using energy properties \cite{FAGCN, zhou2021dirichlet, EEConv}, though they are focused on node classification, where over-smoothing severely impacts performance \cite{2sides}. This narrow focus leaves unexplored how energy dynamics affects other graph tasks. In this work, we provide theoretical and practical insights on leveraging Dirichlet energy to enhance graph classification performance, even in the presence of label noise.

\textbf{Smoothing bias.} 
Most GNNs function as low-pass filters, emphasizing low-frequency components while diminishing high-frequency ones \cite{nt2019revisiting, rusch2023survey}. Specifically, \cite{nt2019revisiting} showed this phenomenon holds for graphs without non-trivial bipartite components, with self-loops further shrinking eigenvalues. Similarly, \cite{oversquashing1} finds that non-bipartite graphs, especially without residual connections, exhibit low-frequency dominance. They also show that continuous-time models like CGNN, GRAND, and PDE-GCND maintain low-pass filtering. \cite{noteonoversmoothing, oversmoothing2} prove that GNN Dirichlet energy exponentially decreases with additional GCN layers when the product of the largest singular value of the weight matrix and the largest eigenvalue of the normalized Laplacian is less than one. Here, it is important to emphasize that \cite{Kang2018RobustGL} examines graph classification under label noise using the mix-up technique. While the mix-up may indirectly promote smoothness in the graph, they do not discuss or establish a relationship between graph smoothness and the Dirichlet energy. Furthermore, their work centers on the smoothness of clusters within the graphs, rather than on the overall smoothness of the graph structure.

\subsection{Lipschitz Continuity in Graph Neural Networks}

Regarding Lipschitz continuity, a key aspect of model robustness, \cite{chuang2022tree} provides a theoretical bound on the Lipschitz constant of the Graph Isomorphism Network (GIN) with respect to the Tree Mover’s Distance (TMD). The derived bound, $|h(G_a)-h(G_b)| \le \sum_{l=1}^{L+1} K^{(l)}_{\phi} \cdot \mathrm{TMD}_w^{L+1}(G_a,G_b)$, relates the change in the GIN's output to the distance between the input graphs as measured by TMD. This theorem highlights that if the constituent learnable functions $\phi^{(l)}$ have bounded Lipschitz constants $K^{(l)}_{\phi}$, then the entire GIN architecture exhibits a Lipschitz property with respect to TMD. Notably, TMD serves as a pseudometric for graphs that are distinguished by the $L$-iteration Weisfeiler-Leman (WL) test, a crucial property given that GIN's representational power is closely tied to the WL test. \cite{davidson2024h} further contribute to the understanding of Lipschitz properties in neural networks operating on sets of features, which are fundamental building blocks in MPNNs. Their analysis of ReLU summation, a common aggregation function, demonstrates that it is uniformly Lipschitz under certain conditions. Moreover, their informal theorem on Hölder MPNN embeddings suggests that if the aggregation, combination, and readout functions within an MPNN are Lipschitz continuous, then the overall MPNN will also be Lipschitz continuous. \cite{juvina2024training} delve into tight Lipschitz constraints for GNNs in the context of node classification. By analyzing a generic graph convolution operation, they derive an optimal Lipschitz constant $\vartheta = \phi(\lambda_K)$ for the network, where $\lambda_K$ is the maximum eigenvalue of a weighted adjacency matrix $M$, assuming non-negative weights and ReLU activations without bias. This work provides a more precise characterization of the robustness of GNNs to input perturbations. \cite{gama2020stability} examine the stability of GNNs with respect to perturbations in the graph shift operator. Their Theorem 4 establishes that if the graph shift operator $S$ is perturbed by $E$ such that $|E|\le\epsilon$, and the filter banks used in the GNN are bounded and the non-linearity is Lipschitz continuous, then the output of the GNN with the perturbed graph $\hat{S}$ will be close to the output with the original graph $S$, with a bound proportional to $\epsilon$ and the number of layers.

\subsection{Oversharpening in Graph Neural Networks}
\label{subsec:oversharpening}
The primary definition of GNN oversharpening, as introduced in the literature and particularly highlighted by analyses such as \cite{GRAFF}, characterizes it as an asymptotic behavior. Specifically, oversharpening occurs when the node features, after passing through multiple GNN layers, become predominantly determined by their projection onto the eigenvector of the graph Laplacian associated with its highest frequency. This implies that the learned representations capture primarily the most rapidly varying components of the signal over the graph.
Pioneering work, notably by \cite{GRAFF}, has rigorously established how the eigenvalues of GNN weight matrices directly influence feature dynamics, leading to either smoothing or sharpening effects. This analysis primarily considers linear graph convolutions employing symmetric weight matrices W.
The key findings are:
\textbf{Positive eigenvalues of W}: These induce an attractive force between the features of connected nodes. This attraction causes their representations to become more similar, promoting a smoothing effect across the graph. Consequently, features tend to align with the low-frequency components of the graph Laplacian, which is characteristic of oversmoothing.
\textbf{Negative eigenvalues of W}: Conversely, these induce a repulsive force between the features of connected nodes. This repulsion drives their representations apart, leading to increased differences and thus a sharpening effect. This enhances the high-frequency components of the features. If these negative eigenvalues are sufficiently dominant and interact appropriately with the graph Laplacian's spectrum, this can lead to the oversharpening phenomenon, where node features become primarily aligned with the highest-frequency eigenvector of the graph Laplacian.
The spectral norm of GNN weight matrices, while not a direct cause of oversharpening in the same way as the sign of eigenvalues, plays a significant modulatory role. It governs the overall "energy" or "scale" of the transformations applied by the GNN layers, thereby influencing the potential for various spectral phenomena, including oversharpening.
The link between a large spectral norm (or large weight variance) and "oversharpening" (defined as high-frequency dominance) is indirect but significant. 
A large spectral norm, by definition, allows for eigenvalues of large magnitudes, both positive and negative.
If learning dynamics or initialization conditions lead to a scenario in which negative eigenvalues of large magnitude become dominant within this expanded spectral envelope, the oversharpening conditions, as described by 
\cite{GRAFF}, could be met.

\cite{zhou2021dirichlet} analyzes this issue through the lens of Dirichlet energy, a measure of the variance of node embeddings. This work shows that the Dirichlet energy at each layer of a Graph Convolutional Network (GCN) is bounded by the Dirichlet energy of the previous layer, scaled by the singular values of the weight matrix. By imposing constraints on the Dirichlet energy, it is possible to control the smoothness of the learned embeddings. The work titled "Graph Neural Networks Do Not Always Oversmooth" challenges the universality of the oversmoothing problem. It establishes a "chaotic, non-oversmoothing phase" in GCNs that can be reached by appropriately tuning the weight variance at initialization. This suggests that oversmoothing is not an inherent limitation of GCN architectures, but rather a consequence of parameter initialization. \cite{eldan2017braess}'s lemma on the spectral gap and edge addition provides insights into how graph structure influences spectral properties, which are related to information propagation and potentially oversmoothing. Their result shows that adding an edge can decrease the spectral gap of the Laplacian matrix under certain conditions related to the eigenvector and degrees of the connected nodes. Finally, the paper \cite{zhuo2024graph} demonstrates that with carefully chosen weights, GNNs can avoid oversmoothing even in deep architectures. Specifically, by employing a whitening transformation on the node features at each layer, the network can prevent the convergence of node representations to a constant vector, suggesting that learnable weights play a crucial role in mitigating oversmoothing.


\section{Dirichlet Energy Amplification in High-Frequency Components Under Label Noise: A Theoretical and Empirical Analysis}
 \label{sec:Amplification}
In continuation of the findings presented in Section~\ref{sec:Dirichlet_meet_GC}, where we established the role of Dirichlet Energy in identifying overfitting in noisy settings, we now deepen this perspective by dissecting the learned representations into frequency components. While we previously observed that elevated Dirichlet Energy in the later phases of training corresponds with the onset of noisy label fitting, our objective here is to uncover which representation subspaces are most impacted, and to understand the underlying dynamics. To this end, we utilize the HLFF-GNN framework~\cite{hlffgnn}, "implemented in our work as \texttt{FGRLConv}",to demonstrate that high-frequency components bear the brunt of overfitting when GNNs are trained on noisy labels.

Empirical results already showed that the total Dirichlet Energy of graph representations tends to rise as the model begins fitting corrupted labels. However, this increase is not uniform across all representation spaces. The HLFF-GNN architecture offers a decomposition into three orthogonal signals: \( Y \) (shared residual), \( Z_1 \) (low-frequency), and \( Z_2 \) (high-frequency). We hypothesize, and confirm, that it is the high frequency subspace \( Z_2 \) that is most vulnerable to label noise. This hypothesis, tested under graph classification (an extension beyond the original node classification setting of HLFF-GNN), is validated both theoretically and empirically.

Consider a graph \( \mathcal{G} = (\mathcal{V}, \mathcal{E}, \mathbf{X}) \), where \( \mathcal{V} \) is the node set, \( \mathcal{E} \subseteq \mathcal{V} \times \mathcal{V} \) the edge set, and \( \mathbf{X} \in \mathbb{R}^{N \times m} \) the input feature matrix. Within HLFF-GNN, node features evolve through frequency modulated propagation into three latent subspaces ~\cite{hlffgnn}. In our \texttt{FGRLConv} implementation:

\begin{itemize}
    \item \( Y \) represents the residual representation,
    \item \( Z_1 \) encodes low-frequency, smooth features propagated via message passing,
    \item \( Z_2 \) captures high-frequency, local features filtered using the graph Laplacian \( \mathbf{\Delta} \).
\end{itemize}

These representations are updated as follows:
\[
Y^{(l+1)} = P - \beta A_Z^{(l)}, \quad
Z_1^{(l+1)} = Z_1^{(l)} - \frac{\beta}{\lambda} A_{YZ_1}^{(l)}, \quad
Z_2^{(l+1)} = \mathbf{\Delta} Z_2^{(l)} - \frac{\beta}{\alpha} A_{YZ_2}^{(l)}
\]

where \( A_{YZ_1}, A_{YZ_2}, A_Z \) are batch aware attention mechanisms modulating signal interactions ~\cite{hlffgnn}. The model minimizes a composite loss:
\[
\mathcal{L} = \|Y - X\|_F^2 + \lambda \operatorname{tr}(Z_1^\top \mathbf{\Delta} Z_1) + \alpha \operatorname{tr}(Z_2^\top (I - \mathbf{\Delta}) Z_2) + \beta (\|Y^\top Z_1\|_F^2 + \|Y^\top Z_2\|_F^2)
\]

Under clean labels, the objective guides the model toward smooth, interpretable feature spaces. However, in the presence of noisy supervision, the model is forced to encode erroneous patterns, disproportionately influencing \( Z_2 \). In the spectral domain, the Dirichlet Energy for \( Z_2 \) becomes:
\[
E^{\text{dir}}(Z_2) = \sum_{r=1}^m \sum_{u=0}^{N-1} \lambda_u (\psi_u^\top Z_{2r})^2
\]

where \( \lambda_u \) and \( \psi_u \) are eigenvalues and eigenvectors of the Laplacian. Larger \( \lambda_u \) correspond to higher frequencies, thereby exaggerating the effect of noise on the \( Z_2 \) energy profile.

\begin{figure}[htbp]
\centering

\begin{subfigure}{0.45\textwidth}
    \includegraphics[width=\linewidth]{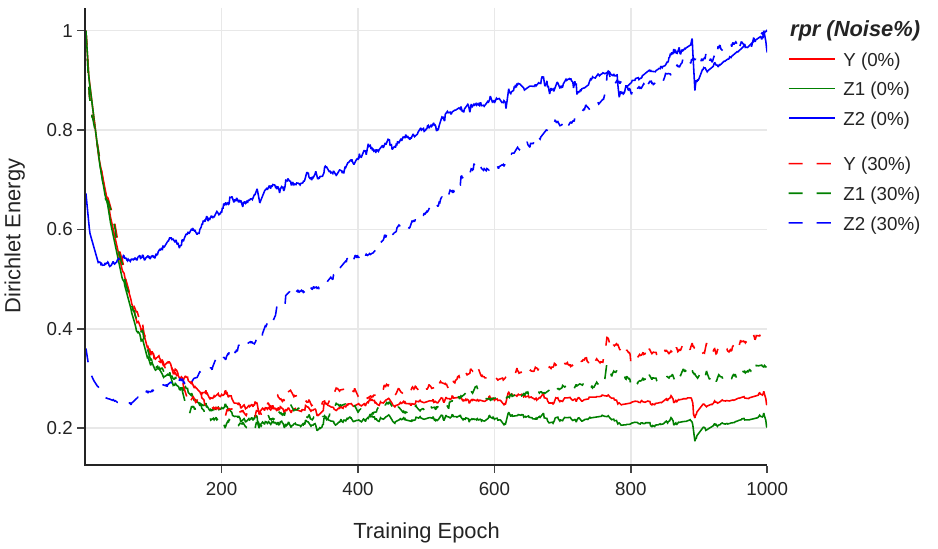}
    \caption{Normalized Dirichlet energy (Y, Z1, Z2)}
\end{subfigure}
\hfill
\begin{subfigure}{0.45\textwidth}
    \includegraphics[width=\linewidth]{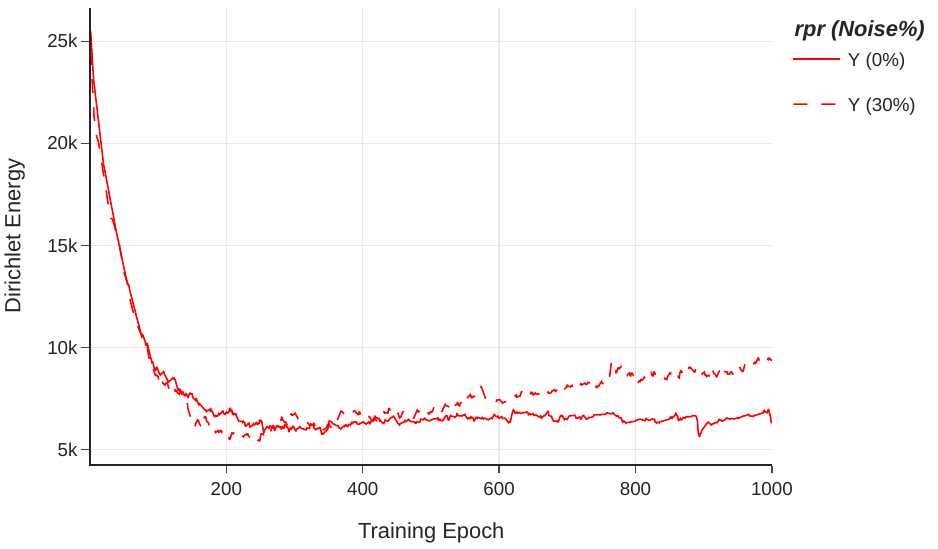}
    \caption{Dirichlet energy of Y}
\end{subfigure}

\vspace{1ex}

\begin{subfigure}{0.45\textwidth}
    \includegraphics[width=\linewidth]{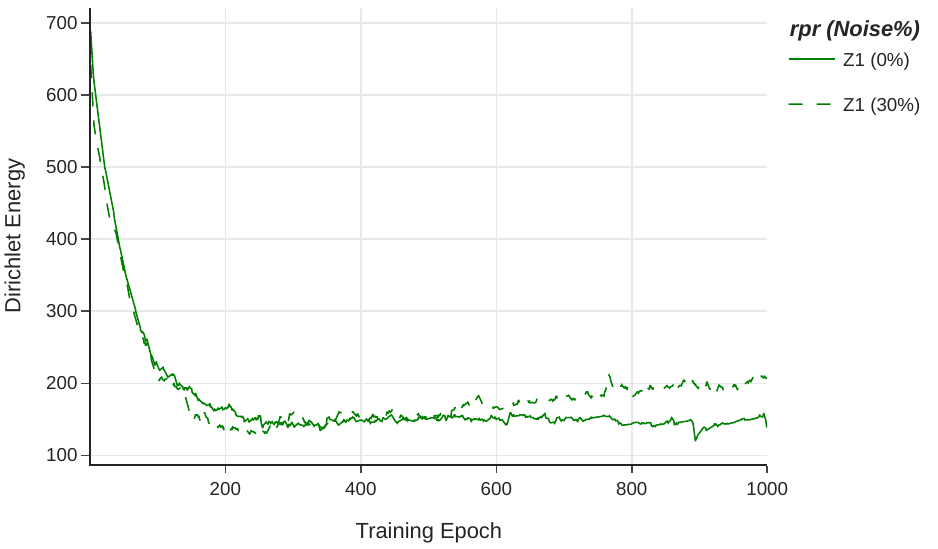}
    \caption{Dirichlet energy of Z1}
\end{subfigure}
\hfill
\begin{subfigure}{0.45\textwidth}
    \includegraphics[width=\linewidth]{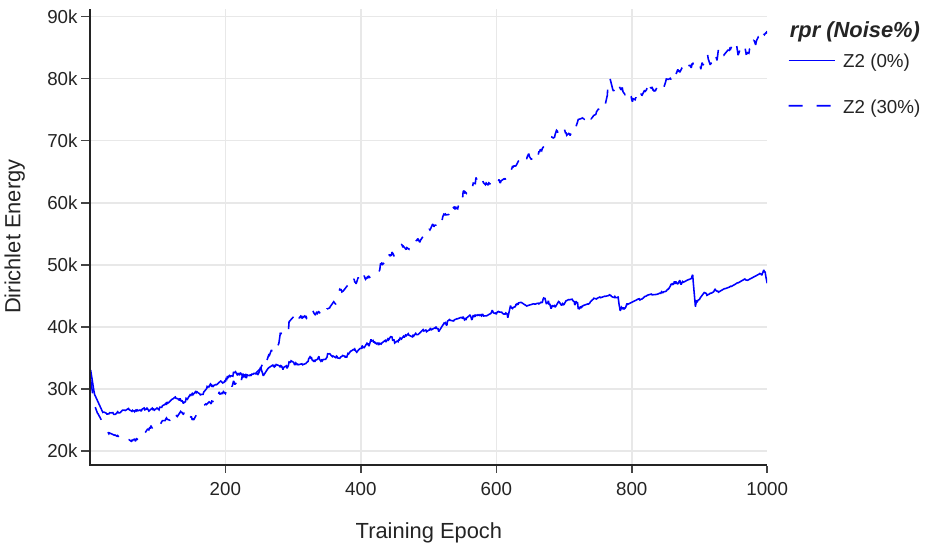}
    \caption{Dirichlet energy of Z2}
\end{subfigure}

\caption{Evolution of Dirichlet energy across training epochs for the representations \( Y \), \( Z_1 \), and \( Z_2 \) learned by the FGRL model on the ENZYMES dataset. Solid lines represent training with 0\% label noise, while dashed lines correspond to 30\% symmetric label noise. The top-left plot shows the normalized energy trajectories for all three representations, with each normalized by its own maximum value to enable direct comparison. The remaining plots display the raw Dirichlet energy for each representation individually, preserving their respective scales to emphasize magnitude differences and noise sensitivity.}
\label{fig:dirichlet_energy_comparison}

\end{figure}

To verify these dynamics, we trained \texttt{FGRLNet} on the ENZYMES dataset under both clean labels and 30\% symmetric label noise. We tracked average per class per sample Dirichlet Energy for \( Y, Z_1, Z_2 \) across training epochs. Under clean supervision, \( Z_2 \)'s energy increased modestly, while \( Z_1 \) and \( Y \) either stabilized or declined. In contrast, noisy supervision triggered a sharp and continuous rise in \( Z_2 \)'s energy, marking it as a reliable signal of overfitting. This phenomenon is visualized in ~\ref{fig:dirichlet_energy_comparison}.

Furthermore, statistical descriptors such as slope and standard deviation of \( E_{\text{dir}}(Z_2) \) were found to be strong early indicators of label noise. Under clean conditions, these metrics remained stable, but they deviated significantly under noisy labels, especially for mislabeled graphs.

These insights lead to actionable strategies for robust training:

\begin{itemize}
    \item High-frequency components (\( Z_2 \)) are principal amplifiers of label inconsistencies.
    \item Monitoring \( E^{\text{dir}}(Z_2) \) dynamics allows early identification of noise driven instability.
    \item Losses can be adaptively modulated to suppress noisy gradient propagation through \( Z_2 \).
\end{itemize}

By grounding robustness in frequency sensitive learning signals, we offer a principled mechanism that can be used to detect and curb overfitting. This analysis extends and reinforces the findings of Section~\ref{sec:Dirichlet_meet_GC}, charting a refined path forward in noise resilient GNN design.

\section{Robust GNN by enforcing positive Eigenvalues of transformations }
\subsection{Proof of Section \ref{sec:positive_eigens}}
\label{app: proof}
\textit{Proof of Proposition} \ref{theorem: avg_dirichlet}. Let us denote $\Lambda = \{\lambda_u^i, 0 \leq u \leq N^i \wedge 1 \leq i \leq n\}$ as the set of the all graph frequencies in $\mathcal{D}$ and we rewrite it as $\Lambda = \{\lambda_k | 0 \leq k \leq N_{tot} \wedge N_{tot} = \sum_{i = 0}^{n} N^i\}$. This formalization is agnostic to the specific graph in the dataset. \\
From this we can easily rewrite \eqref{eq: spectral_avg_dirichlet_onegraph} as follows:
\begin{equation}
\label{eq: spectral_avg_dirichlet_onegraph}
   E^{dir}(\mathcal{D}) := \frac{1}{|\mathcal{D}|} \sum_{r = 1}^{m} \sum_{u = 1}^{N_{tot}} \lambda_u(\bm{\psi}_u^{\top} \mathbf{Z}^{tot}_{r})^2,
\end{equation}
having $\mathbf{Z}^{tot} \in \mathbb{R}^{N_{tot} \times m'}$ and $\bm{\psi}_u \in \mathbb{R}^{N_{tot} \times 1}$. \\
Let us assume now the case of a graph $\mathcal{G} = (\mathbf{Z}, \mathbf{A})$, where $\mathbf{Z} = [\mathbf{Z^1} \|\ldots \| \mathbf{Z^n}] \in \mathbb{R}^{N_{tot} \times m'}$, and $\mathbf{A} \in \mathbb{R}^{N_{tot} \times N_{tot}}$ is a diagonal block matrix, where each block $i$ in the diagonal is $\mathbf{A}^i$. The resulting $E^{dir}(\mathbf{Z})$ can be computed as: 
\begin{equation}
\label{eq: spectral_dirichlet_onegraph}
   E^{dir}(\mathbf{Z}) := \sum_{r = 1}^{m} \sum_{u = 1}^{N_{tot}} \lambda_u'(\bm{\psi}_u'^{\top} \mathbf{Z}^{tot}_{r})^2,
\end{equation}
Let's notice that \eqref{eq: spectral_dirichlet_onegraph} differs from \eqref{eq: spectral_avg_dirichlet_onegraph} in their set of eigenvalues and eigenvectors, and the scaling factor $|D|$.\\
Let us now define the graph Laplacian of $\mathcal{G}$ as $\mathbf{\Delta} = \mathbf{I_{N_{tot}}} - \mathbf{D}^{-\frac{1}{2}}\mathbf{A}\mathbf{D}^{-\frac{1}{2}}$. Being $\mathcal{G}$ composed by disconnected graphs we can write its Laplacian as the following diagonal block matrix:
\begin{equation}
\label{eq: unique_laplacian}
\mathbf{\Delta} = \begin{bmatrix}
\mathbf{I}_{N^1} - (\mathbf{D}^1)^{-\frac{1}{2}}\mathbf{A}^1(\mathbf{D}^1)^{-\frac{1}{2}} & \cdots & 0 \\
\vdots & \ddots & \vdots \\
0 & \cdots & \mathbf{I}_{N^n} - (\mathbf{D}^n)^{-\frac{1}{2}}\mathbf{A}^n(\mathbf{D}^n)^{-\frac{1}{2}}
\end{bmatrix} = \begin{bmatrix}
\mathbf{\Delta}^1 & \cdots & 0 \\
\vdots & \ddots & \vdots \\
0 & \cdots & \mathbf{\Delta}^n
\end{bmatrix}
\end{equation}
From \ref{eq: unique_laplacian}, we can evince that the eigenvalues set of eigenvalues $\Lambda'$ of $\mathbf{\Delta}$ corresponds to the union of the eigenvalues for each Laplacian of the disconnected graphs s.t. $\Lambda' = \{\Lambda^i| 1 \leq i \leq n\}$. This derives from the property that $det(\Delta - \lambda\mathbf{I}_{N_{tot}}) = \prod_{i = 1}^n det(\Delta^i - \lambda\mathbf{I}_{N^i})$ \cite{linear_algebra}. So this proves that $\lambda_u \equiv \lambda_u', \forall u$ in Equations \ref{eq: spectral_avg_dirichlet_onegraph} and \ref{eq: spectral_dirichlet_onegraph}. 

For the eigenvectors, suppose $\bm{\psi}_j^i$ is the $j$-th eigenvector of $\mathbf{\Delta}^i$ corresponding to eigenvalue \( \lambda_j^i \) (e.g. $j \in \{0, \ldots, N^i\}$). We construct the corresponding eigenvector of $\mathbf{\Delta}$ through the diagonal block matrix properties. Formally, the corresponding eigenvector $\bm{\psi}_u'$ of $\mathbf{\Delta}$ corresponding to \( \lambda_j^i \) is given by:
\[
\bm{\psi}_u' = \begin{bmatrix}
0 \\
\vdots \\
0 \\
\bm{\psi}_j^{i} \\
0 \\
\vdots \\
0
\end{bmatrix} = \begin{bmatrix}
0 \\
\vdots \\
0 \\
\bm{\psi}_u \\
0 \\
\vdots \\
0
\end{bmatrix}
\]
This vector $\bm{\psi}_u'$ satisfies the eigenvector equation for $\mathbf{\Delta}$:
\[
\mathbf{\Delta} \bm{\psi}_u' = \lambda_u \bm{\psi}_u' = \lambda'_u \bm{\psi}_u'
\]
From this, it follows always that $\bm{\psi}_u^{\top} \mathbf{Z}^{tot}_{r} = \bm{\psi}_j^{i\top} \mathbf{Z}^{i}_{r}, \forall r$. \\Thus, it follows that $E^{dir}(\mathcal{D}) = |\mathcal{D}| \cdot E^{dir}(\mathbf{Z})$.

\section{Additional Empirical Studies on Dirichlet Energy Behaviour}
\label{sec:Additional_studies}

To gain deeper insight into the behavior of the Dirichlet energy during training, we conducted two additional experiments. These studies aim to clarify how energy evolves under different training dynamic, specifically in scenarios of overfitting and varying levels of label noise.

\textbf{Overfitting on clean data}

\begin{figure}[htbp]
\centering

\begin{subfigure}{0.45\textwidth}
    \includegraphics[width=\linewidth]{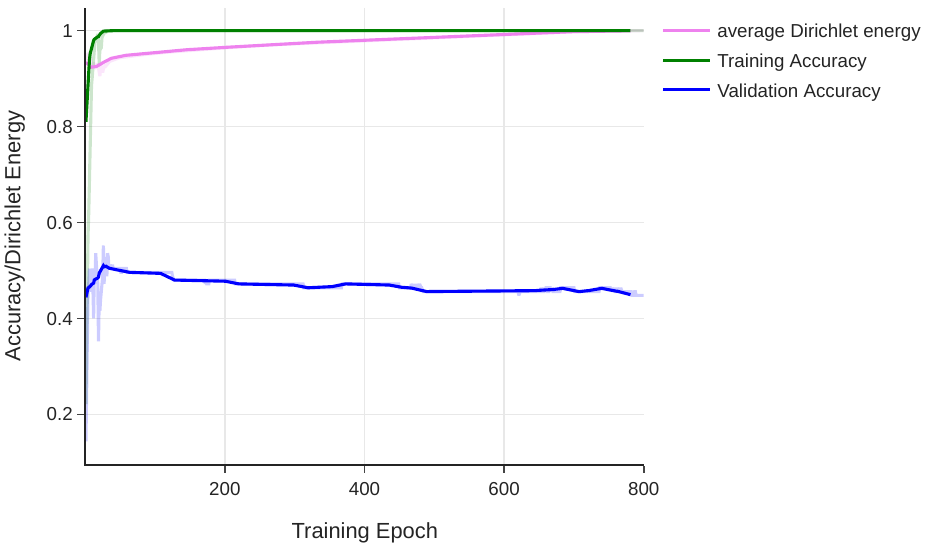}
    \caption{}
    \label{fig:enzymes_energy}
\end{subfigure}
\hfill
\begin{subfigure}{0.45\textwidth}
    \includegraphics[width=\linewidth]{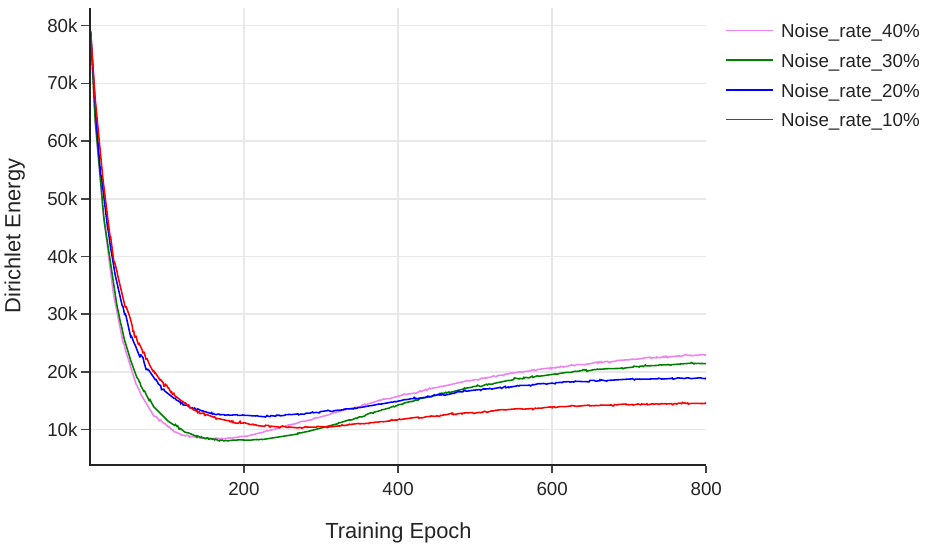}
    \caption{}
    \label{fig:noise_levels_energy}
\end{subfigure}

\caption{Empirical observations on Dirichlet energy dynamics. (a) Training on ENZYMES without noise, where the model is deliberately overfitted. The plot shows normalized Dirichlet energy alongside training and validation accuracy. As the model memorizes the training data, Dirichlet energy increases, indicating a rise in high-frequency components. (b) Evolution of normalized Dirichlet energy during training on PPA with symmetric label noise levels ($10\%$ to $40\%$). All curves follow a similar pattern: an initial energy decrease followed by a rise. Higher noise levels result in higher final energy, suggesting a link between Dirichlet energy growth and the amount of noisy labels.}
\label{fig:Additional_experiments}

\end{figure}

The first experiment investigates the evolution of Dirichlet energy when a model is intentionally overfitted to clean data.
We trained a GIN model on the ENZYMES dataset without regularization and with the explicit goal of fitting the training data completely. As shown in Figure~\ref{fig:enzymes_energy}, the model successfully overfits the training set, as evidenced by the near-perfect training accuracy and the large gap between training and validation accuracy.

Notably, the Dirichlet energy consistently increases throughout the training process. This finding suggests that an upward trend in energy is not necessarily caused by label noise, but may instead be a general result of overfitting. In particular, the model’s growing capacity to memorize fine-grained details may lead to less smooth and more fluctuating feature representations, reflected by higher Dirichlet energy.

\textbf{Training under varying levels of noise}

The second experiment investigates how the Dirichlet energy evolves during training when the dataset contains varying levels of label noise.
A GIN model was trained on the PPA dataset under symmetric label noise at rates of $10\%, 20\%, 30\%,$ and $40\%$. During training, we tracked the evolution of the Dirichlet energy according to the noise rate.

As illustrated in Figure~\ref{fig:noise_levels_energy}, all noise levels exhibit a similar pattern in energy evolution: an initial decrease followed by a rise. This U-shaped trajectory suggests that the model initially learns generalizable low-frequency patterns, then begins to memorize label noise, resulting in less smooth node representations and thus higher Dirichlet energy.

Crucially, we observe that higher label noise levels consistently lead to higher final Dirichlet energy. The $40\%$ noise curve ends with the highest energy, while the $10\%$ noise setting maintains the lowest. This trend highlights a direct relationship between label noise and energy growth, further suggesting that Dirichlet energy can serve as an indicator of the extent to which the model is fitting noise.

\clearpage
\section{Additional Tables}
\begin{table}[htbp]
  \caption{Sensitivity analysis of $lr_u$ for $u$ under 40\% asymmetric noise, $\% \textbf{Change}$ shows the percentage difference of test Accuracy using GCOD with two different learning rates for $u$.}
    \label{tab:lr_sensitivity}
    \centering
    \begin{tabular}{lccc}
        \hline
        \textbf{Dataset} & \textbf{$lr_u=1$} & \textbf{$lr_u=0.1$} & \textbf{\% Change} \\
        \hline
        Proteins    & 76.19 & 75.89 & -0.39\% \\
        MNIST       & 72.61 & 72.48 & -0.18\% \\
        Enzymes     & 69.81 & 68.33 & -2.12\% \\
        IMDB/Binary & 72.89 & 71.94 & -1.30\% \\
        Mutag       & 93.19 & 92.61 & -0.62\% \\
        Reddit      & 47.94 & 48.08 & +0.29\% \\
        MSRC/21     & 95.57 & 95.45 & -0.13\% \\
        \hline
    \end{tabular}
\end{table}
\begin{table}[htbp]
        \caption{Relative runtime comparison of GIN trained with standard Cross Entropy (baseline, runtime normalized to 1.00) versus alternative robustness-enhancing methods (SOP, GCOD, and CE+W2) on the PPA dataset (using 30\% data, 6 classes).}
        \label{tab:runtime_comparison}
        \centering
        \begin{tabular}{lc}
            \hline
            \textbf{Method} & \textbf{Runtime} \\
            \hline
            GIN  & 1.00 \\
            SOP  & 1.048 \\
            GCOD & 1.029 \\
            CE+W2 & 1.33 \\
            \hline
        \end{tabular}
\end{table}
\begin{table}[htbp]
\centering
    \caption{Performance of CE vs.\ GCOD with 20 \% symmetric label noise.
      The last column reports the difference (GCOD – CE) on test accuracy.}
    \label{tab:alldataset}

    \begin{adjustbox}{width=0.55\linewidth}
      \begin{tabular}{llcccc}
        \hline
        \textbf{Dataset} & \textbf{Metric} &
        \textbf{0 \% CE} & \textbf{20 \% CE} & \textbf{20 \% GCOD} &
        \makecell[c]{20 \%\\(GCOD–CE)} \\ \hline

        \multirow{3}{*}{MNIST}
          & Best       & 72.69 & 66.30 & 71.26 & +4.96 \\
          & Last       & 69.80 & 53.64 & 67.88 & +14.24 \\ \cline{3-5}
          & Difference & 2.89  & 12.66 & 3.38  &        \\ \hline

        \multirow{3}{*}{ENZYMES}
          & Best       & 73.33 & 64.16 & 68.69 & +4.53 \\
          & Last       & 65.78 & 57.50 & 62.54 & +5.04 \\ \cline{3-5}
          & Difference & 7.55  & 6.66  & 6.15  &        \\ \hline

        \multirow{3}{*}{MSRC\_21}
          & Best       & 96.69 & 90.26 & 94.69 & +4.43 \\
          & Last       & 93.80 & 79.64 & 90.26 & +10.62 \\ \cline{3-5}
          & Difference & 2.89  & 10.62 & 4.43  &        \\ \hline

        \multirow{3}{*}{PROTEINS}
          & Best       & 81.16 & 76.23 & 79.38 & +3.15 \\
          & Last       & 79.18 & 62.32 & 78.12 & +15.80 \\ \cline{3-5}
          & Difference & 1.98  & 13.91 & 1.26  &        \\ \hline

        \multirow{3}{*}{MUTAG}
          & Best       & 94.73 & 89.47 & 90.01 & +0.54 \\
          & Last       & 84.21 & 68.42 & 86.84 & +18.42 \\ \cline{3-5}
          & Difference & 10.52 & 21.05 & 3.17 &        \\ \hline

        \multirow{3}{*}{\makecell[l]{IMDB-\\BINARY}}
          & Best       & 76.50 & 75.00 & 75.40 & +0.40 \\
          & Last       & 71.60 & 71.00 & 73.50 & +2.50 \\ \cline{3-5}
          & Difference & 4.90  & 4.00  & 1.90  &        \\ \hline

        \multirow{3}{*}{\makecell[l]{REDDIT-\\MULTI-12K}}
          & Best       & 48.15 & 45.05 & 44.98 & –0.07 \\
          & Last       & 46.01 & 44.67 & 44.89 & +0.22 \\ \cline{3-5}
          & Difference & 2.14  & 0.38  & 0.09  &        \\ \hline
      \end{tabular}
    \end{adjustbox}
\end{table}


\section{Additional Figures}
\label{appendix:additional_figures}

\begin{figure}[ht]
    \centering
    \includegraphics[page=1,width=.5\textwidth]{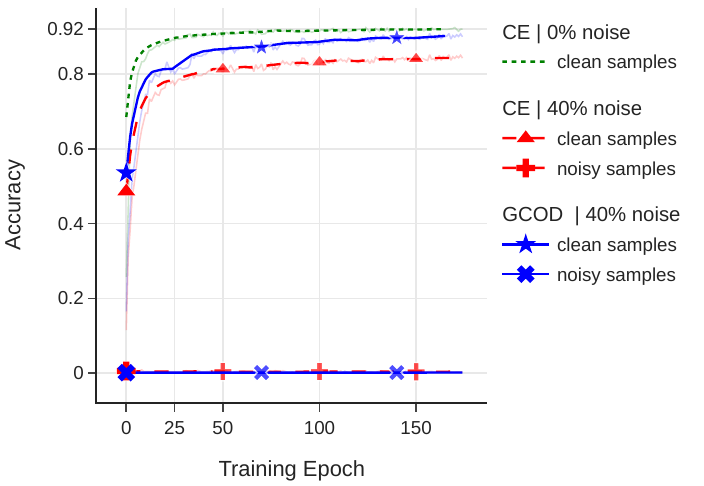}
    \caption{Training accuracy for known noisy and clean samples using GCN with CE loss. (4 class form PPA, with 40\% symmetrical noise)
    }
    \label{fig:noisy_pure_40_1}
\end{figure}

\begin{figure}[ht]
    \centering   
    \includegraphics[page=1,width=.5\textwidth]{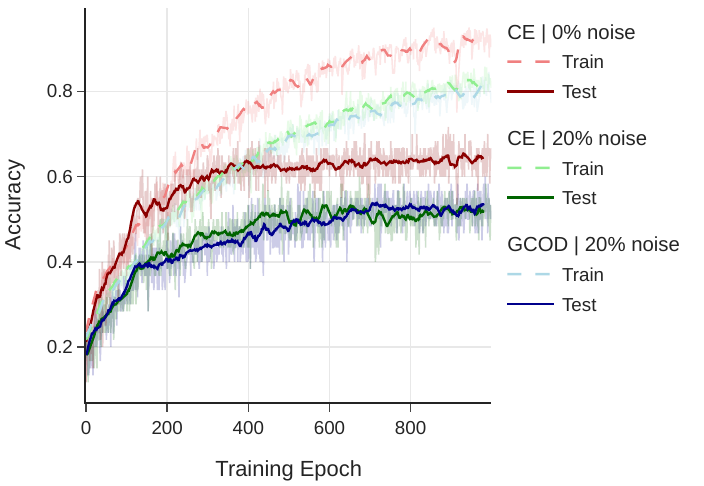}
    \caption{Comparison of the train and test accuracy for the Enzymes dataset. GCN model with different losses and noise levels.  }
    \label{fig:NCOD_accuracy}
\end{figure}

\begin{figure}[ht]
    \centering   
    \includegraphics[page=1,width=0.5\textwidth]{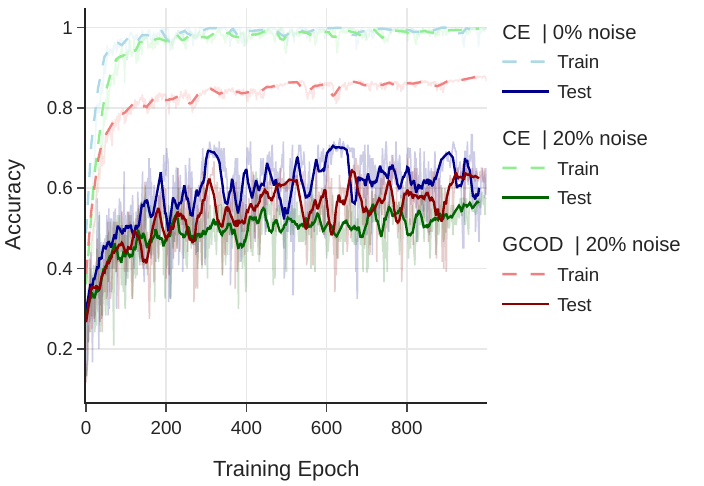}
    \caption{Comparison of the train and test accuracy for the Enzymes dataset with GIN model, on clean and 20\% symmetric noise.}
    \label{fig:gin_enzymes}
\end{figure}

\begin{figure}[ht]
    \centering   
    \includegraphics[page=1,width=0.5\textwidth]{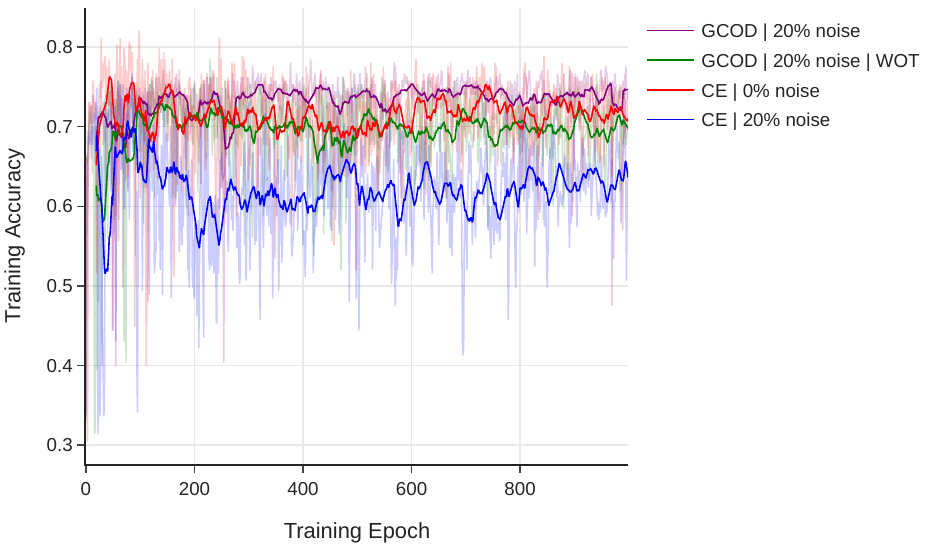}
    \caption{Ablation study showing that scaling $u$ by training accuracy prevents it from growing too aggressively. 
    Without this scaling (GCOD WOT), $u$ dominates early and harms generalization, whereas in GCOD where $u$ is scaled by training accuracy activates $u$ gradually and achieves higher test accuracy.}
    \label{fig:uablation}
\end{figure}

\begin{figure}[ht]
    \centering   
    \includegraphics[page=1,width=.5\textwidth]{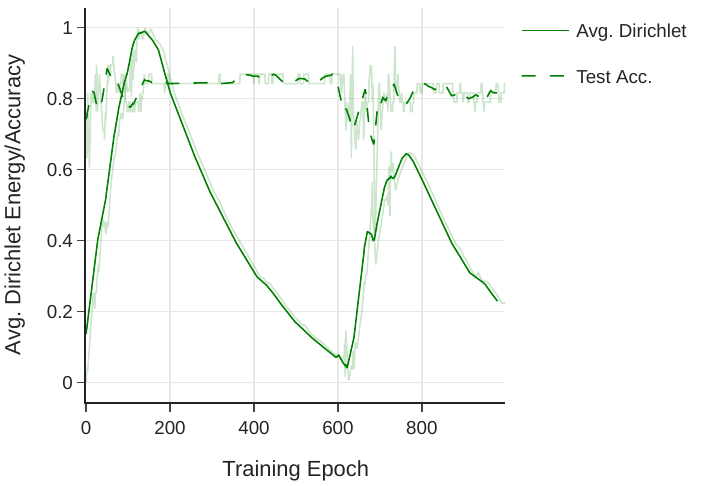}
    \caption{Average test accuracy and average Dirichlet energy on the MUTAG dataset with 0\% label noise using the GIN model. The plot illustrates the evolution of accuracy and representation smoothness over training epochs.}
    \label{fig:mutag}
\end{figure}

\begin{figure}[ht]
    \centering   
    \includegraphics[page=1,width=0.5\textwidth]{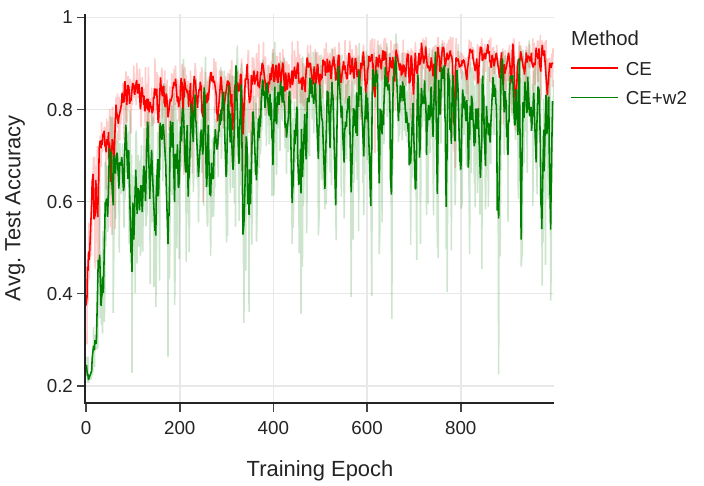}
    \caption{Test accuracy on the PPA dataset (30\% subset, 6-class task) using Cross-Entropy (CE) and the CE+W2 method, which enforces positive-semidefinite weight matrices via eigendecomposition. 
    While comparable peak accuracy, CE+W2 exhibits unstable convergence due to the spectra constraint applied after each epoch disrupts optimization. 
    }
    \label{fig:unstability}
\end{figure}


\end{document}